\documentclass{article}

\usepackage[preprint]{tmlr}

\usepackage[utf8]{inputenc} 
\usepackage[T1]{fontenc}    
\usepackage{hyperref}       
\usepackage{url}            
\usepackage{booktabs}       
\usepackage{amsfonts}       
\usepackage{nicefrac}       
\usepackage{microtype}      
\usepackage{xcolor}         
\usepackage{wrapfig}
\usepackage{diagbox}
\usepackage{venndiagram}
\usepackage{tikz}

\usepackage{amsmath,amsfonts,amsthm,bm,mathtools}

\usepackage{multirow}
\usepackage[utf8]{inputenc} 
\usepackage[T1]{fontenc}    
\usepackage{hyperref}       
\usepackage{url}            
\usepackage{booktabs}       
\usepackage{amsfonts}       
\usepackage{nicefrac}       
\usepackage{microtype}      
\usepackage{amsmath}
\usepackage{algorithm}
\usepackage{graphicx}
\usepackage{adjustbox}
\usepackage{amssymb}
\usepackage{breqn}
\usepackage[noend]{algpseudocode}
\usepackage{caption}

\def\eqref#1{equation~\ref{#1}}

\def\1{\bm{1}}

\DeclareMathAlphabet{\mathsfit}{\encodingdefault}{\sfdefault}{m}{sl}
\SetMathAlphabet{\mathsfit}{bold}{\encodingdefault}{\sfdefault}{bx}{n}

\DeclareMathOperator*{\argmax}{argmax}

\newcommand{\numnew}{\ensuremath{m}}
\newcommand{\numfromhist}{\ensuremath{m^{(h)}}}
\newcommand{\finallabelset}{\ensuremath{\mathfrak{L}}}

\def\month{11}  
\def\year{2023} 
\def\openreview{\url{https://openreview.net/forum?id=T55dLSgsEf}} 

\title{Accelerating Batch Active Learning Using Continual Learning Techniques}

\author{%
Arnav Das$ {\thanks{Equal contribution}}~~^\dagger$ \hspace{0.25em} Gantavya Bhatt$^{*\dagger}$  \hspace{0.25em} Megh Bhalerao$^{\dagger}$ \hspace{0.25em} Vianne Gao$^{\diamond}$ \hspace{0.25em} Rui Yang$^{\diamond}$ \hspace{0.25em} Jeff Bilmes$^{\dagger}$ \\
$^{\dagger} $University of Washington, Seattle \quad $^\diamond$ Memorial Sloan Kettering Cancer Center \\
\texttt{\{arnavmd2, gbhatt2, bilmes\}@uw.edu} \\
}

\begin{document}

\maketitle

\begin{abstract}
  A major problem with Active Learning (AL) is high training costs
  since models are typically retrained from scratch after every query
  round. We start by demonstrating that standard AL on neural networks
  with warm starting fails, both to accelerate training and to
  avoid catastrophic forgetting when using fine-tuning over AL query rounds.  
  We then develop a new class
  of techniques, circumventing this problem, by biasing further
  training towards previously labeled sets. We accomplish this by employing
  existing, and developing novel, replay-based Continual Learning (CL)
  algorithms that are effective at quickly learning the new without
  forgetting the old, especially when data comes from an evolving
  distribution. We call this
  paradigm \textit{"Continual Active
  Learning" (CAL)}.  We show CAL achieves significant speedups using
  a plethora of replay schemes that use model distillation and that
  select diverse/uncertain points from the history.
  We conduct
  experiments across many data domains, including natural language,
  vision, medical imaging, and computational biology, each with
  different neural architectures and dataset sizes. CAL
  consistently provides a $\sim$3x reduction in training time, while retaining performance and out-of-distribution robustness, showing its wide applicability.\looseness-1
\end{abstract}

\vspace{-1\baselineskip}
\section{Introduction}
\label{sec: intro}
\vspace{-0.1\baselineskip}


While neural networks have been successful in a variety of different supervised settings, most such approaches are labeled-data hungry and require significant computation. From a large pool of unlabeled data, active learning (AL) selects subsets of points to label by imparting the learner with the ability to query a human annotator. Such methods incrementally add points to the labeled pool by repeatedly: (1) training a model from scratch on the current labeled pool and (2) using some measure of model uncertainty and/or diversity to select a set of points to query the annotator~\citep{settles2009active, settles2010, fass2015, badge2020, glister2021}. AL has been shown to reduce the amount of training data
required but can be computationally expensive since it requires retraining a model, typically from scratch, after each query round.\looseness-1


A \textit{simple} solution is to warm start the model parameters between query rounds.
However, the observed speedups tend to still be limited since the model must make several passes through an ever-increasing pool of data. Moreover, warm starting alone in some cases can hurt generalization, as discussed in \cite{Ash2020} and \cite{distil2021}. Another extension to this is to solely train on the newly labeled batch of examples to avoid re-initialization. However, as we show in Section~\ref{Active Learning as Continual Learning}, naive fine-tuning fails to retain accuracy on previously seen examples since the distribution of the query pool may drastically change with each round.\looseness-1

This problem of \textit{catastrophic forgetting} while incrementally learning from a series of new tasks with shifting distributions is a central question in another paradigm called Continual Learning (CL) \citep{french1999catastrophic,mccloskey1989catastrophic,mcclelland1995there,kirkpatrick2017overcoming}. CL has recently gained popularity, and many algorithms have been introduced to allow models to quickly adapt to new tasks without forgetting~\citep{MER2018, GEM2017, AGEM2019, aljundi2019gradient, chaudhry2020using, ewc2017}.

In this work, we propose Continual Active Learning (CAL)\footnote{There have been papers published with titles containing the ``Continual Active Learning'' phrase, but these do not merge {\bf Continual} Learning with {\bf Active Learning} as we do, hence our name.}, which applies Continual Learning strategies to accelerate batch Active Learning. In CAL, we apply CL to enable the model to learn the newly labeled points without forgetting previously labeled points while using past samples efficiently using  \textit{replay-based} methods. As such, we observe that CAL attains significant training time speedups over standard AL. This is beneficial for the following reasons: {\bf (1):}
As neural networks swell~\citep{megatron}, so do the environmental and financial model training
costs~\citep{stochasticparrot2021, Dhar2020, greenai2020}. Reducing the number of gradient updates required for AL will help mitigate such costs, especially with large-scale models.
{\bf (2):}
Reducing AL computational requirements makes AL more accessible for edge computing, IoT, and low-resource device deployment~\citep{edgeAL} such as with federated learning~\citep{li2020federated}.
{\bf (3):}
Developing new AL algorithms/acquisition functions, or searching for architectures as done with NAS/AutoML that are well-suited \emph{specifically} for AL, can require hundreds or even thousands of runs. Since CAL's speedups are agnostic to the AL algorithm and the neural architecture, such experiments can be significantly sped up.
Overall, the importance of speeding machine learning training processes is well recognized, as evidenced by the plethora of efforts in the computing systems community~\citep{unity2022, slaq2017, alpa2022}.\looseness-1


\begin{wrapfigure}[20]{r}{0.5\textwidth}
    \vskip -0.1in	
    \centering	
    \includegraphics[width=0.44\textwidth]{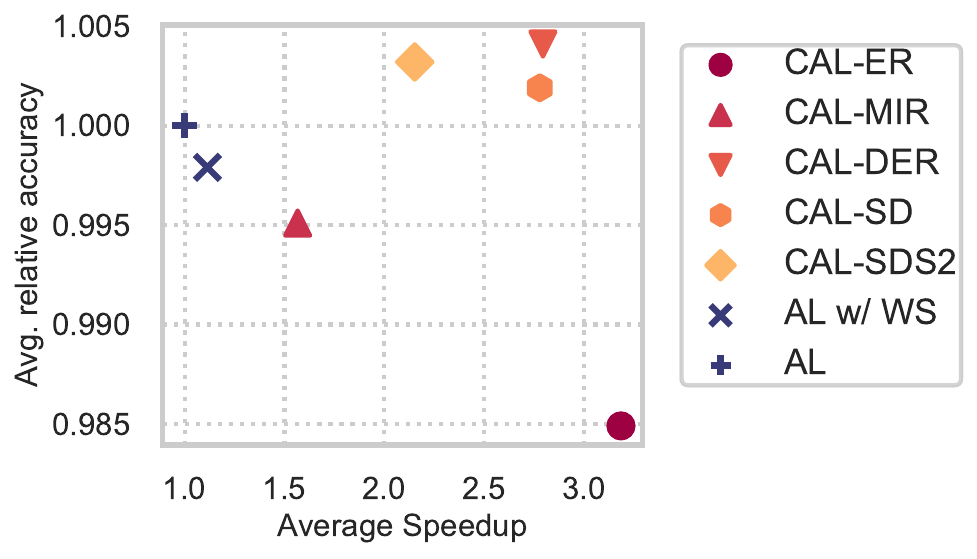}
    \caption{\small Summary of our results: Benchmarking average relative accuracy vs.\ speedup for the proposed methods (\emph{solid markers}) against the baseline. Each point is averaged across datasets and labeling budgets; \emph{top-right region is desiderata}. This shows that proposed CAL methods (particularly CAL-DER/SD/SDS2) have equivalent performance relative to the baseline (if not better) on multiple datasets and across different budgets, all while providing significant speedups.}
    \label{fig:main_result}	
\end{wrapfigure}



In addition, CAL demonstrates another practical application for CL methods. Many settings used to benchmark CL methods in recent work are somewhat contrived \citep{farquhar2018towards, wallingford2023fluid}. Most CL work considers the class/domain incremental setting, where only the samples that belong to a subset of the set of classes/domains of the original dataset are available to the model at any given time.
This setting need not be the only benchmark upon which CL methods are evaluated. We suggest that the evaluation of future new CL algorithms should be determined not only on traditional CL evaluation schemes and benchmarks but also on their performance in the CAL setting.

In the present work, we are not using CL to improve or change AL querying strategies. We view this as both a strength of the present work and an opportunity for future work. Firstly, it is a strength of our present work since any AL query strategy, both old and new, can in principle be applied in the CAL setting, both for speeding up the AL strategy and offering, as argued above, a test bed for CL. Indeed, a major
AL research challenge includes combining AL with other techniques, as we believe we have done herein.  Secondly, it is an opportunity since there is nothing inherent in the definition of CL that precludes the CL tasks from being dependent on the model as we show below.  There may be a way for CL to open doors to intrinsically new AL querying policies of attaining new batches of unlabeled data. This we leave to the future.  Our present {\em core goal is to accelerate batch AL training via CL techniques while preserving accuracy}.


To the best of our knowledge, this application of CL algorithms to accelerate batch AL has never been explored. Our contributions can be summarized as follows:
{\bf (1)} We first demonstrate that batch active learning techniques can benefit from continual learning techniques, and their merger creates a new class of techniques that we propose to call the ``CAL framework.'' {\bf (2) } We benchmark several existing CL methods (CAL-ER, CAL-DER, CAL-MIR) as well as novel methods (CAL-SD, CAL-SDS2) and evaluate them on diverse datasets based on the accuracy/speedup they can attain over standard AL. {\bf (3) } We study speedup/performance trade-offs on datasets that vary in modality (natural language, vision, medical imaging, and computational biology), neural architecture with varying degrees of computation (Transformers/CNNs/MLPs), data scale (including some larger datasets, one having 2M samples), and class-balance.
And {\bf (4)}, lastly, we demonstrate that models trained with CAL and standard AL models behave similarly, in that both classes of models attain similar uncertainty scores on held-out datasets and achieve similar robustness performance on out-of-distribution data.
Figure~\ref{fig:main_result} summarizes our results, detailed later in the paper and greatly detailed in the appendices.\looseness-1



\vspace{-1\baselineskip}
\section{Related Work}
\vspace{-.2\baselineskip}
Active learning~\citep{atlas1989training,CohnAL94, fass2015, glister2021,badge2020} has demonstrated label efficiency over passive learning. In addition, there has been extensive work on theoretical aspects of AL~\citep{guillory2009-actnoncon,hanneke2009theoretical, 10.1145/1273496.1273541, Balcan2010} where~\cite{JMLR:v13:hanneke12a} shows sample complexity advantages over passive learning in noise-free classifier learning for VC classes. More recently, Active Learning has also been studied as a procedure to incrementally learn the underlying data distribution with the help of discrepancy framework \citep{mathelin2022discrepancybased, discre, discreAISTATS}.  \looseness-1

Recently there has been an interest in speeding up active learning since most deep learning
is computationally demanding. \cite{NEURIPS2019_95323660,NEURIPS2019_84c2d486,sener2018active} aim to reduce the number of query iterations by having large query batch sizes. However, they do not exploit the learned models from previous rounds for the subsequent ones and are therefore complementary to CAL. Work such as \cite{DBLP:journals/corr/abs-2007-00077, 10.1145/1321440.1321461, 9093556, DBLP:journals/corr/ZhuB17, zhang2023labelbench} speeds up the selection of the new query set by appropriately restricting the search space or by using generative methods. This work can be easily integrated into our framework because CAL works on the training side of active learning, not on the query selection. On the other hand, \cite{LEWIS1994148, Coleman2020Selection, DBLP:journals/corr/abs-1905-03677} use a smaller proxy model to reduce computation overhead, however, they still follow the standard active learning protocol, and therefore can be accelerated when integrated with CAL. 


There exists work that explores continual/transfer learning and active learning in the same context. \cite{Perkonigg2021-ee} propose an approach that allows active learning to be applied to data streams
of medical images by introducing a module that detects domain shifts.  This is quite distinct from our work: our work uses CL algorithms to prevent catastrophic forgetting and to accelerate learning. \cite{ZHOU2021101997} study when
standard active learning is used to fine-tune a pre-trained model, and employs transfer learning --- this does not consider continual learning and active learning together, however, and is therefore not related to our work. Finally,~\cite{FoCAL2022} studies where a robot observes unlabeled data sampled from a shifting distribution, but does not explore active learning acceleration.

For preventing catastrophic forgetting, we mostly focus on replay-based algorithms that are state-of-the-art methods in CL. However, as demonstrated in Section~\ref{Active Learning as Continual Learning} on how active learning rounds can be viewed in a continual learning context, one can apply other methods such as EWC~\citep{doi:10.1073/pnas.1611835114}, Bayesian divergence priors~\cite{li2007bayesian}, structural regularization~\citep{DBLP:journals/corr/abs-2104-08604} or functional regularization~\citep{Titsias2020Functional} as well.

The effect of warm-started model training on generalization and convergence speed has been explored by \cite{Ash2020} which empirically demonstrates that a model that has been pretrained on a source dataset converges faster but exhibits worse generalization on a target dataset when compared to a randomly initialized model. However, that work only considers the setting where the source and target datasets are unbiased estimates of the same distribution. This is distinct from our work since the distributions we consider are all dependent on the model at each AL round. Furthermore, our work employs CL methods in addition to warm-starting, also not considered in~\cite{Ash2020}.

\section{Background}
\label{sec:background}

\subsection{Batch Active Learning}
\begin{figure*}[h]
    \centering
    \includegraphics[trim={0 0 0 0}, clip, width=\linewidth]{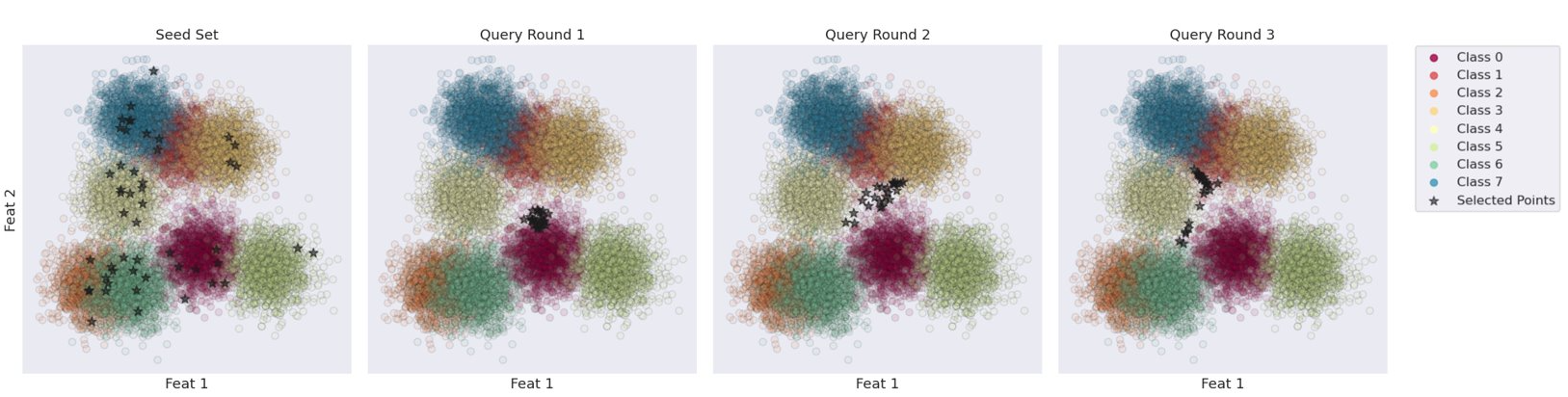} 
    \caption{\small Visualization of standard AL when training a linear model on a simple 2D synthetic dataset. Each plot shows which samples were chosen by uncertainty sampling to be added to the labeled pool at given a query round. Notably, the distributions of the new labeled points vary significantly between rounds motivating CAL. \looseness-1}
    \label{fig:AL_vis}
    \vspace{-.7\baselineskip}
\end{figure*} 

Define $[n] = \{1, ..., n\}$, and let $\mathcal{X}$ and $\mathcal{Y}$ denote the input and output domains respectively. AL typically starts with an unlabeled dataset $\mathcal{U} = \{x_i\}_{i \in [n]}$, where each $x_i \in \mathcal{X}$. The AL setting allows the model $f$, with parameters $\theta$, to query a user for labels for any $x \in \mathcal{U}$, but the total number of labels is limited to a budget $b$, where $b \leq n$. Throughout the work, we consider classification tasks so the output of $f(x; \theta)$ is a probability distribution over classes. The goal of AL is to ensure that $f$ can attain low error when trained only on the set of $b$ labeled points.

\vspace{-.8\baselineskip}

\noindent \\ Algorithm~\ref{alg:active_learning_loop} details the general AL procedure. Lines 3-6 construct the seed set $\mathcal{D}_1$ by randomly sampling a subset of points from $\mathcal{U}$ and labeling them. Lines 7-14 iteratively expand the labeled set for $T$ rounds by training the model from a random initialization on $\mathcal{D}_t$ until convergence and selecting $b_t$ points (where $\sum_{t \in [T]} b_t = b$) from $\mathcal{U}$ based on some selection criteria that is dependent on $\theta_t$. The selection criteria generally select samples based on model uncertainty and/or diversity~\citep{lewis1994, dagan1995, settles2012, glister2021, fass2015, badge2020, https://doi.org/10.48550/arxiv.1708.00489}. In this work, we primarily consider uncertainty sampling \cite{lewis1994, dagan1995, settles2012}, though we also test other selection criteria in Section~\ref{sec:appen_other_acquisition_func} in the Appendix.

\begin{wrapfigure}[13]{r}{0.6\textwidth}
    \begin{minipage}{0.6\textwidth}
    \vskip -0.4in	
    \centering	
    \begin{algorithm}[H]
    \label{alg:AL_procedure}
    \small
    \begin{algorithmic}[1]
    \vspace{-.1\baselineskip}
    \Procedure{ActiveLearning}{$f$, $\mathcal{U}$, $b_{1:T}$, $T$}
        \State $t \leftarrow 1$, $\finallabelset \leftarrow \emptyset$ \Comment{{\tiny Initialize}}
        \State $\mathcal{U}_t \sim \mathcal{U}$ \Comment{{\tiny Draw $b_1$ samples from $\mathcal{U}$}}
        \State $\mathcal{D}_t \leftarrow \{(x_i,y_i) | x_i \in \mathcal{U}_t\} $ \Comment{{\tiny Provide labels}}
        \State $\mathcal{U} \leftarrow \mathcal{U} \setminus \mathcal{U}_t$ \Comment{{\tiny Remove from unlabeled set}}
        \State $\finallabelset \leftarrow \finallabelset \cup \mathcal{D}_t $ \Comment{{\tiny Add to labeled set}}
        \While{$t \leq T$}  
            \State Randomly initialize $\theta_{init}$
            \State $\theta_t \leftarrow \text{Train}(f, \theta_{init}, \finallabelset) $
            \State $\mathcal{U}_{t+1} \leftarrow \text{Select}(f, \theta_t, \mathcal{U}, b_t)$  \Comment{{\tiny Select $b_t$ points based on $\theta_t$}}
            \State $\mathcal{D}_{t+1} \leftarrow \{(x_i,y_i) | x_i \in \mathcal{U}_t\} $
            \State $\mathcal{U} \leftarrow \mathcal{U} \setminus \mathcal{U}_{t+1}$; $\finallabelset \leftarrow \finallabelset \cup \mathcal{D}_{t+1} $; $t \leftarrow t + 1$
        \EndWhile  
        \State $\theta_T \leftarrow \text{Train}(f, \theta_{init}, \finallabelset) $
        \State return $\finallabelset$, $\theta_T$
    \EndProcedure
    \caption{\label{alg:active_learning_loop} Batch Active Learning}
    \end{algorithmic}
    \end{algorithm}
    \end{minipage}
\end{wrapfigure}

\vspace{-.5\baselineskip}
\paragraph{Uncertainty Sampling} is a widely-used practical AL method that selects
$\mathcal{U}_t = \{x_1, \dots, x_{b_t}\} \subseteq \mathcal{U}$ by choosing the samples that maximize a notion of model uncertainty. We consider entropy \citep{dagan1995} as the uncertainty metric, so if $h_\theta(x) \triangleq -\sum_{i \in [k]} f(x;\theta)_i \log f(x;\theta)_i$, then $\mathcal{U}_{t+1} \in \underset{\mathcal A \subset \mathcal{U} : |\mathcal A| = b_t}{\text{argmax}} \sum_{x \in \mathcal A} h_{\theta_t}(x)$.

\paragraph{Distribution Shift} As shown in Algorithm~\ref{alg:active_learning_loop}, the set of samples that are labeled at a given query round are dependent on the model parameters. Since the model is repeatedly updated, the distribution of newly labeled samples may vary across different AL rounds. This is illustrated in Figure~\ref{fig:AL_vis}, where we perform standard AL in a simplified setting. In the example, we use a linear model to perform classification on a 2D synthetic dataset and perform AL with entropy-based uncertainty sampling as the acquisition function. Upon visualizing the set of samples that are queried by the model at different AL rounds, it is evident that distribution shift does occur. We demonstrate empirically in Figure~\ref{fig:model_structure} that training on newly labeled samples alone will cause catastrophic forgetting due to distribution shifts.

\vspace{-.5\baselineskip}
\subsection{Continual Learning}
We define $\mathcal{D}_{1:n} = \bigcup_{i \in [n]} \mathcal{D}_i$. In CL, the  dataset consists of $T$ tasks $\{\mathcal{D}_1,...,\mathcal{D}_T\}$ that are presented to the model sequentially, where $\mathcal{D}_t = \{(x_i, y_i)\}_{i \in N_t}$, $N_t$ are the task-$t$ sample indices, and $n_t=|N_t|$. At time $t \in [T]$, the data/label pairs are sampled from the current task $(x,y) \sim \mathcal{D}_t$, and the model has only limited access to the history $\mathcal{D}_{1:t-1}$.
The CL objective is to efficiently adapt the model to $\mathcal{D}_t$ while ensuring performance on the history does not appreciably degrade. 
We focus on replay-based CL techniques that attempt to approximately solve CL optimization by using samples from $\mathcal{D}_{1:t-1}$ to regularize the model while adapting to $\mathcal{D}_{t}$. Please refer to appendix~\ref{sec:appen_CL} for more details on CL.



\begin{wrapfigure}[10]{r}{0.6\textwidth}
    \begin{minipage}{0.6\textwidth}
    \vspace{-1\baselineskip}	
    \centering	
    \small
    \begin{algorithm}[H]
    \small
    \label{alg:CL_procedure}
    \begin{algorithmic}[1]
    \vspace{-.1\baselineskip}
    \Procedure{ContinualTrain}{$f$, $\theta_0$, $\mathcal{D}$, $\mathcal{H}$, $\numnew$, $\numfromhist$}
        \State $\tau \leftarrow 0$ \Comment{{\tiny $\tau$ is a local iteration index.}}
        \While{not converged}  
            \State  $\tau \leftarrow \tau + 1$ 
            \State $\mathcal{B}_{\text{current}} \leftarrow \{(x_i,y_i)\}_{i=1}^m \sim \mathcal{D}$ \Comment{{\tiny Sample from new task.}}
            \State $\mathcal{B}_{\text{replay}} \leftarrow \text{Select}(f, \theta_{\tau-1}, \mathcal{H}, \numfromhist)$ \Comment{{\tiny Sample from history.}}
            \State $\theta_\tau \leftarrow \text{Update}(f, \theta_{\tau-1}, \mathcal{B}_{\text{current}}, \mathcal{B}_{\text{replay}})$
        \EndWhile  
        \State return $\theta_\tau$
    \EndProcedure
    \vspace{-.2\baselineskip}
    \caption{\label{alg:continual_learning} Continual Learning}
    \end{algorithmic}
    \end{algorithm}
    \end{minipage}

\end{wrapfigure}

    \noindent Algorithm~\ref{alg:continual_learning} outlines general replay-based CL, where the objective is to adapt $f$ parameterized by $\theta_0$ to $\mathcal{D}$ while using samples from the history $\mathcal{H}$. $\mathcal{B}_{\text{current}}$ consists of $\numnew$ points randomly sampled from the \emph{current time's} $\mathcal{D}$, and $\mathcal{B}_{\text{replay}}$ consists of $\numfromhist$ points chosen based on a
criterion that selects from $\mathcal{H}$.
In line 6, $\theta_{\tau}$ is computed based on an update rule that utilizes both $\mathcal{B}_{\text{replay}}$ and $\mathcal{B}_{\text{current}}$. 

Different CL methods assume varying degrees of access to the history. In many practical settings, $\mathcal{D}_{1:T}$ is too large to store in memory or $T$ is unknown so CL algorithms assume limited or no access to samples from the history. Some of these approaches remove unimportant samples from the history \cite{aljundi2019gradient, aser2020}, while others solely rely on regularizing the model parameters without replaying any samples from the history \cite{ewc2017, lwf2017}. In the CAL setting, $\mathcal{D}_{1:T}$ is already stored in memory as is done in the standard AL setting so employing a CL algorithm that imposes a memory constraint would needlessly cause CAL to underperform. In Table~\ref{tab:CL_methods}, a handful of prior CL works are sorted by what assumptions about memory/history access they make; methods that do not make any memory constraint assumption are the most well-suited for CAL. A more comprehensive and complete overview of CL methods can be found in \cite{CLSurvey2022}.



\begin{table}[]
\centering
\begin{tabular}{|c|ccc|}
\toprule
\multirow{2}{*}{\textbf{Continual Learning Papers}}              & \multicolumn{3}{c|}{\textbf{History Access}} \\
                         & \textbf{None} & \textbf{Partial} & \textbf{Full} \\
\hline
\cite{ewc2017}, \cite{lwf2017}            &  $\checkmark$ &  $\times$ & $\times$ \\
\hline
\cite{aljundi2019gradient}, \cite{aser2020} &  $\times$ &  $\checkmark$ & $\times$ \\
\hline
\cite{DER2020}, \cite{mir2019} & & & \\
\cite{GEM2017}, \cite{AGEM2019} &  $\times$ & $\checkmark$ & $\checkmark$ \\
\cite{Ratcliff1990} & & & \\
\bottomrule
\end{tabular}
\caption{Different previous CL work organized by the memory access assumptions made in their settings. In the CAL setting, we do not need to assume that we have limited access to the history so we only consider CL methods that are applicable to the setting where the history is fully accessible. This category primarily contains replay-based methods.}
\label{tab:CL_methods}
\end{table}

\vspace{-.5\baselineskip}
\section{Blending Continual and Active Learning}
\label{sec:will-the-blend}
\label{Active Learning as Continual Learning}
\vspace{-.5\baselineskip}



\begin{figure*}[h]
    \centering
    \includegraphics[trim={0 0 0 0}, clip, width=0.99\linewidth]{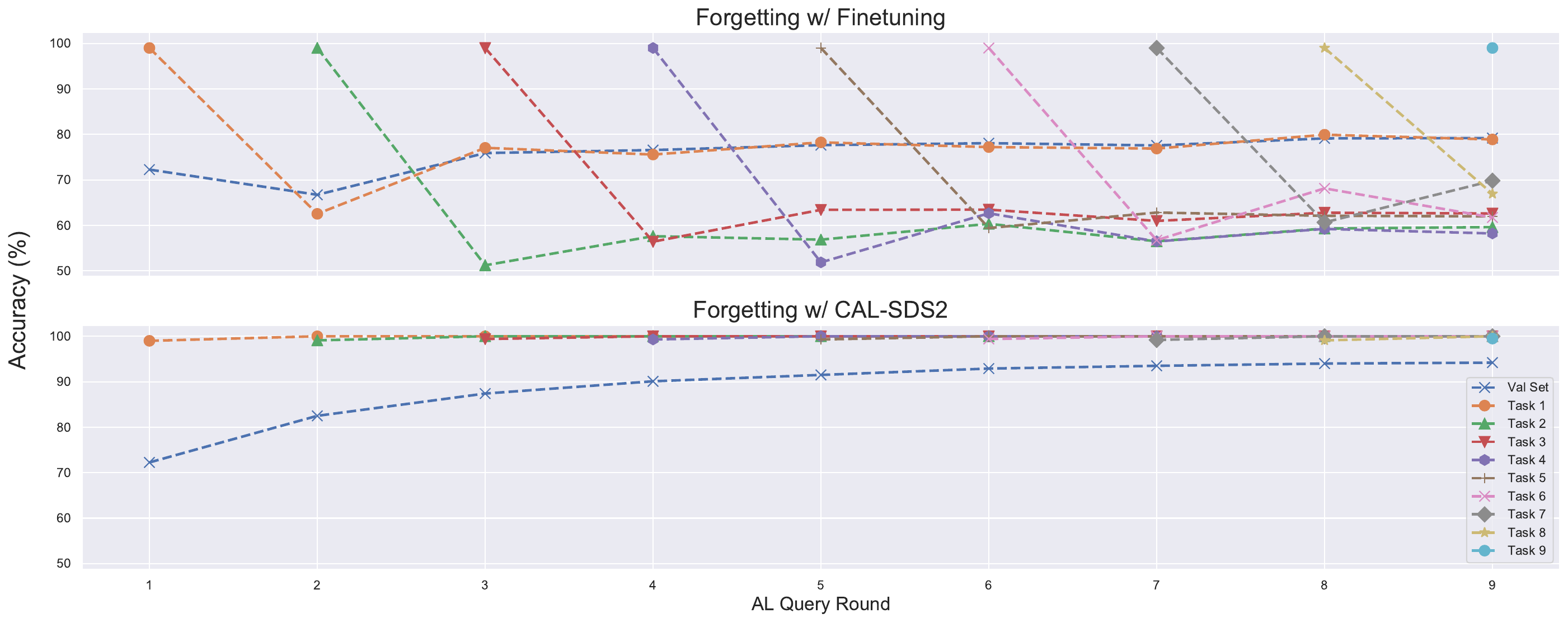} 
    \caption{\small A ResNet-18's performance on CIFAR-10 on the validation
      set (blue X) and task $i$ for $i \in \{1,2,\dots, 9\}$ (other
      colors) after training the warm-started model and training just
      on the new task data (top) compared to efficiently adapting to newly labeled data with a CAL method (bottom).  In the top plot, at each round a new 5\% of
      the full dataset is added to the labeled pool and the previously
      trained model is further trained {\em just} on the new points
      (the isolated upper-right cyan point shows task 9's performance
      after round 9). After each round, the accuracy on the
      corresponding task does well, but all previous tasks'
      accuracies drop precipitously, demonstrating a form of
      catastrophic forgetting.  In the bottom plot, the same setting is considered but the model is adapted to the new data with CAL-SDS2, one of the CAL methods. Also, the validation set performance
      has flatlined despite continued training on new data when finetuning, but continues to increase when CAL-SDS2 is employed. This
      demonstrates that naively
      fine-tuning a model to newly labeled points can work poorly.\looseness-1}
    \label{fig:model_structure}
    \vspace{-.7\baselineskip}
\end{figure*}

A clear AL inefficiency is that the model $f$ is retrained from
scratch on the entire labeled pool after every query round. One
potential solution idea is to simply continue training the model only
on the newly AL-queried samples and, via the process of warm starting,
hope that history will not fade.  Unfortunately for this approach,
Figure~\ref{fig:model_structure} (top) shows that when the model is
warm-start trained \emph{only} on the task $t$ (samples labeled at AL
round $t$ using entropy sampling), historical sample performance
deteriorates precipitously while performance on the validation set
flatlines. That is, at AL round $t$ (x-axis), we continue to train the
model until convergence on task $t$ and track accuracy (y-axis) on
each previous task and also on the validation set.  The performance of
task 1, after the initial drop, tracks that of the validation set,
since task 1 is a model agnostic initial unbiased random subset query
of the training data. The performance of task $i$, for $i>1$, however,
each of which is the result of model-conditioned AL query, shows
perilous historical forgetting.  In the end, the model performs
considerably worse on all of the historical tasks (aside from task 1)
than on the validation set, even though it has been trained on those
tasks and not on the validation set.  This experiment suggests that: (1)
the distribution of each AL-queried task $t > 1$ is different than the
data distribution; (2) fine-tuning to task $t$ can result in
catastrophic forgetting; and (3) techniques to combat catastrophic
forgetting are necessary to effectively incorporate new information
between successive AL rounds.


\vskip 0.08in
The CAL approach, shown in Algorithm~\ref{alg:CAL_procedure}, uses CL
techniques to ameliorate catastrophic forgetting.  The key difference
between CAL and Algorithm~\ref{alg:active_learning_loop} is line
9. Instead of standard training, replay-based CL is used to adapt $f$
to $\mathcal{D}_t$ while retaining performance on
$\mathcal{D}_{1:t-1}$. The speedup comes from two sources: (1) we
are computing gradient updates only on a useful subset of the history
$\mathcal{D}_{1:t-1}$ rather than all of it
for reasonable choices of $\numfromhist$; and (2) the model converges faster since it
starts warm. The ability of CAL to combat catastrophic forgetting is shown in Figure Figure~\ref{fig:model_structure} (bottom).
In the rest of the section, define
$\mathcal{L}_{c}(\theta) \triangleq \mathop{\mathbb{E}}_{(x, y) \sim
  \mathcal{B}_{\text{current}}} \left [ \ell (y, f\left(x; \,
    \theta)\right) \right]$.  We next define and compare several CAL
methods and assess their performance based on their performance on the
test set and the speedup they attain compared to standard AL.

\begin{wrapfigure}[16]{r}{0.6\textwidth}
    \begin{minipage}{0.6\textwidth}
    \vspace{-1.5\baselineskip}
    \centering	
    \begin{algorithm}[H]
    \small
    \begin{algorithmic}[1]
    \Procedure{CAL}{$f$, $\mathcal{U}$, $b_{1:T}$, $T$, $\numnew$, $\numfromhist$}
        \State $t \leftarrow 1$, $\finallabelset \leftarrow \emptyset$ \Comment{{\tiny Initialize}}
        \State $\mathcal{U}_t \sim \mathcal{U}$ \Comment{{{\tiny Draw $b_1$ samples from $\mathcal{U}$}}}
        \State $\mathcal{D}_t \leftarrow \{(x_i,y_i) | x_i \in \mathcal{U}_t\} $ \Comment{{\tiny Provide labels}}
        \State $\mathcal{U} \leftarrow \mathcal{U} \setminus \mathcal{U}_t$
        \State $\finallabelset \leftarrow \finallabelset \cup \mathcal{D}_t $
        \State Randomly initialize $\theta_{0}$
        \While{$t \leq T$}  
            \State \textcolor{blue}{$\theta_t$ $\leftarrow$ ContinualTrain($f$, $\theta_{t-1}$, $\mathcal{D}_t$, $\mathcal{D}_{1:t-1}$, $\numnew$, $\numfromhist$)}
            \State $\mathcal{U}_{t+1} \leftarrow \text{Select}(f, \theta_t, \mathcal{U}, b_t)$ \Comment{{\tiny Select $b_t$ points from $\mathcal{U}$}}
            \State $\mathcal{D}_{t+1} \leftarrow \{(x_i,y_i) | x_i \in \mathcal{U}_{t+1}\} $
            \State $\mathcal{U} \leftarrow \mathcal{U} \setminus \mathcal{U}_{t+1}$; $\finallabelset \leftarrow \finallabelset \cup \mathcal{D}_{t+1} $; $t \leftarrow t + 1$
        \EndWhile 
        \State $\theta_T$ $\leftarrow$ ContinualTrain($f$, $\theta_{T-1}$, $\mathcal{D}_T$, $\mathcal{D}_{1:T-1}$, $\numnew$, $\numfromhist$)
        \State return $\finallabelset$, $\theta_T$
    \EndProcedure
    \caption{\label{alg:CAL_procedure} The general CAL approach.}
    \end{algorithmic}
    \end{algorithm}
    \end{minipage}
\end{wrapfigure}

\vspace{-.5\baselineskip}

\paragraph{Experience Replay (CAL-ER)} is the simplest and oldest replay-based method \citep{Ratcliff1990, Robins95}. In this approach, $\mathcal{B}_{\text{current}}$ and $\mathcal{B}_{\text{replay}}$ are interleaved to create a minibatch $\mathcal{B}$ of size $\numnew + \numfromhist$ and $\mathcal{B}_{\text{replay}}$ is chosen uniformly at random from $\mathcal{D}_{1:t-1}$. The parameters $\theta$ of model $f$ are updated based on the gradient computed on $\mathcal{B}$.

\vspace{-.5\baselineskip}
\paragraph{Maximally Interfered Retrieval (CAL-MIR)} chooses a size-$\numfromhist$ subset of points
from $\mathcal{D}_{1:t-1}$
most likely to be forgotten \citep{mir2019}. Given a batch of $\numnew$ labeled samples $\mathcal{B}_{\text{current}}$ sampled from $\mathcal{D}_t$ and model parameters $\theta$, $\theta_v$ is computed by taking a ``virtual'' gradient step i.e., $\theta_v = \theta - \eta \nabla \mathcal{L}_c(\theta) $ where $\eta$ is the learning rate. Then for every example $x$ in the history, $s_{MIR}(x) = \ell (f(x; \theta), y) - \ell (f(x; \theta_v), y)$ (i.e., the change in loss after taking a single gradient step) is computed. The $\numfromhist$ samples with the highest $s_{MIR}$ score are selected for $\mathcal{B}_{\text{replay}}$, and the remainder is similar to experience replay. $\mathcal{B}_{\text{current}}$ and  $\mathcal{B}_{\text{replay}}$ are concatenated to form the minibatch (as in CAL-ER), upon which the gradient update is computed. In practice, selection is done on a random subset of $\mathcal{D}_{1:t-1}$ for speed, since computing $s_{MIR}$ for every historical sample is prohibitively expensive.\looseness-1

\vspace{-.5\baselineskip}
\paragraph{Dark Experience Replay (CAL-DER)} uses a distillation approach to regularize updates~\citep{DER2020}. Let $g(x;\theta)$ denote the pre-softmax logits of classifier $f(x;\theta)$, i.e., $f(x; \theta)  =  \text{softmax}(g(x;\theta))$. In DER, every $x' \in \mathcal{D}_{1:t-1}$ has an associated $z'$ which corresponds to the model's logits at the end of the task when $x$ was first observed --- if $x' \in \mathcal{D}_{t'}$, then $z' \triangleq g(x'; \theta_{t'}^*))$ where $t' \in [t-1]$ and $\theta_{t'}^*$ are the parameters obtained after round $t'$. DER minimizes $\mathcal{L}_{\text{DER}}(\theta)$ defined as:
\begin{align}
    \begin{split}
      \mathcal{L}_{\text{DER}}(\theta) \triangleq \mathcal{L}_{c}(\theta)
      &+ \mathop{\mathbb{E}}_{(x',y', z') \sim \mathcal{B}_{\text{replay}}} \left[ \alpha\, \|g(x';\theta) - z' \|_2^2 \left.+ \right. \beta \, \ell(y',f(x';\theta)) \right], \raisetag{\baselineskip}
    \end{split}
\end{align}%
where $\mathcal{B}_{\text{replay}}$ is a batch uniformly at randomly (w/o replacement) sampled from $\mathcal{D}_{1:t-1}$, and $\alpha$ and $\beta$ are tuneable hyperparameters. The first term ensures that samples from the current task are classified correctly. The second term consists of a classification loss and a mean squared error (MSE) based distillation loss applied to historical samples.


\paragraph{Scaled Distillation (CAL-SD)} $\mathcal{L}_{\text{SD}}(\theta)$ is a new objective proposed in this work defined via:
\begin{align}
    \begin{split}
    \mathcal{L}_{\text{replay}}(\theta) \triangleq &
    \mathop{\mathbb{E}}_{(x',y',z') \sim \mathcal{B}_{\text{replay}}} \biggl[ \alpha \,  D_{\text{KL}}\Bigl( \text{softmax}(z') \mid \mid f(x' ; \theta) \Bigr) + (1 - \alpha)  \, \ell \bigl( y', f(x';\theta) \bigr) \biggr], \raisetag{\baselineskip}
    \end{split}
\end{align}%
and then
$\mathcal{L}_{\text{SD}}(\theta) \triangleq \lambda_t \,
\mathcal{L}_{c}(\theta) + (1 - \lambda_t) \,
\mathcal{L}_{\text{replay}}(\theta)$ where
$\lambda_t \triangleq |\mathcal{D}_t|/\bigl(|\mathcal{D}_t| +
|{D}_{1:t-1}|\bigr)$.  Similar to CAL-DER,
$\mathcal{L}_{\text{replay}}$ is a sum of two terms: a distillation
loss and a classification loss. The distillation loss expresses the KL-divergence between the
posterior probabilities produced by $f$ and $\text{softmax}(z')$,
where $z'$ is defined in the DER section. We use KL-divergence
instead of MSE loss on the logits so that the distillation and
the classification losses have the same scale and dynamic
range, and also since it allows the logits to drift by a constant term that does not affect the softmax output but does effect the MSE loss. $\alpha \in [0,1]$ is a tuneable hyperparameter.\looseness-1

The weight of each term is determined adaptively by a
``stability/plasticity'' trade-off term $\lambda_t$.  A
stability-plasticity dilemma is commonly found in both biological and
artificial neural networks \citep{Abraham2005, stabilityplasticity}. A
network is \emph{stable} if it can effectively retain past information
but cannot adapt to new tasks efficiently, whereas a network that is
\emph{plastic} can quickly learn new tasks but is prone to
forgetting. The trade-off between stability and plasticity is a
well-known constraint in CL~\citep{stabilityplasticity}. For CAL, we
want the model to be plastic early on, and stable later on. We apply
this intuition with $\lambda_t$: higher values indicate higher
plasticity, since minimizing the classification error of samples from
the current task is prioritized. Since $\mathcal{D}_{1:t-1}$ increases
with $t$, $\lambda_t$ decreases and the model becomes more stable in
later training rounds.


\vspace{-.5\baselineskip}
\paragraph{Scaled Distillation w/ Submodular Sampling (CAL-SDS2)}
CAL-SDS2 is another new CL approach we introduce in this work. CAL-SDS2 uses CAL-SD to regularize the model and uses a submodular sampling procedure to select a diverse representative set of history points to replay. Submodular functions are well-known to be suited to capture notions of diversity and representativeness~\citep{lin-bilmes-2011-class, fass2015, submodular2022} and the simple greedy algorithm can approximately maximize, under a cardinality constraint, a monotone submodular function up to a $1 - e^{-1}$ constant factor multiplicative guarantee~\citep{fisher1978analysis, minoux1978accelerated, mirzasoleiman2015lazier}. We define our submodular function $G$ as:
\begin{align}
    G(\mathcal{S}) \triangleq \sum_{x_i \in \mathcal{A}} \max_{x_j \in \mathcal{S}} w_{ij} + \lambda \log{ \left ( 1 + \sum_{x_i \in \mathcal{S}} h(x_i) \right )}.
\end{align}
The first term is a facility location function, where $w_{ij}$ is a similarity score between samples $x_i$ and $x_j$. We use $w_{ij} = \exp{(-\|z_i - z_j \|^2 / 2 \sigma^2)}$ where $z_i$ is the penultimate layer of model $f$ for $x_i$ and $\sigma$ is a hyperparameter. The second term is a concave over modular function~\citep{featurebased2013} and $h(x_i)$ is a standard AL measure of model uncertainty, such as entropy of the model's output distribution. Both terms are well known to be monotone non-decreasing submodular, as is their non-negatively weighted sum~\citep{submodular2022}.
The core reason for applying a concave function over a model-uncertainty-score-based modular function, instead of keeping it as a pure modular function, is
to better align the modular uncertainty values with the facility location function. Otherwise, if
we do not apply the concave function,
the facility location function dominates during the early steps of the greedy algorithm and the modular function
dominates in the later steps of greedy.
In order to speed up SDS2 (to avoid the need to perform forward passes over the entire history before a single step), we randomly subsample the history indices
to produce $\mathcal A$ and re-compute forward passes to produce fresh $z_i$ values for $i \in \mathcal A$ before re-constructing $G(\mathcal S)$, and we then perform submodular maximization; thus $\mathcal{S} \subset \mathcal{A} \subset \mathcal{D}_{1:t-1}$. 
The objective of CAL-SDS2 is to ensure that the set of samples that are replayed are both difficult and diverse, similar to the motivation of the heuristic employed in~\cite{fass2015}.

\paragraph{Baselines} All proposed CAL methods are compared against two AL baselines. The first baseline is standard active learning, denoted as AL, which is no different from the procedure shown in Algorithm~\ref{alg:active_learning_loop}. We also consider active learning with warm starting (AL w/ WS), which uses the converged model from the previous round to initialize the model for the current round. Both models retrain on the entire dataset at each query round.

\section{Experiments and Results}
\label{sec:experiments-results}
\vspace{-.5\baselineskip}

\begin{figure*}[tbh]
\centering
\includegraphics[width=\textwidth]{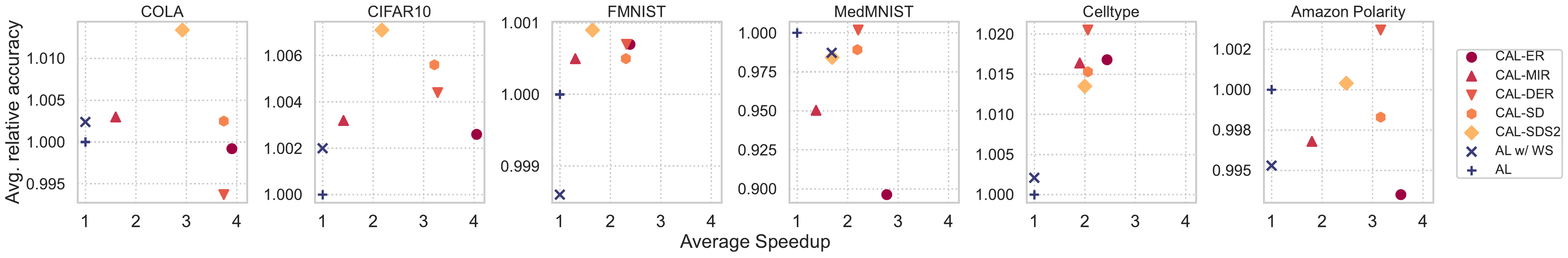}
\captionsetup{width=\textwidth}
\caption{\small Relative accuracy vs.\ speedup averaged over different
  labeling budgets. For every dataset, we have \emph{at least one} CAL
  method, that is faster with same accuracy than
  baseline (\emph{top right region is desiderata}).}
\label{fig:all_datasets}
\vspace{-1\baselineskip}
\end{figure*}

We evaluate the validation performance of the model when we train on different fractions ($b/n$) of the full dataset. We compute the {\em speedup} attained by a CAL method by dividing the wall-clock training time of the
baseline AL method over the wall-clock training time of the CAL method. We test the CAL methods on a variety of different datasets spanning multiple modalities. Two baselines do not utilize CAL which are standard {\em AL} (active learning) as well as {\em AL w/ WS} (active learning with warm starting but still training using all the presently available labeled data). 

Our objective is to demonstrate: (1) \emph{at least one} CAL-based method exists that can match or outperform a standard active learning technique while achieving a significant speedup for every budget and dataset and (2) models that have been trained using a CAL-based method behave no differently than standard models. We emphasize that the purpose of this work is \emph{not} to champion a single method, but rather to showcase an assortment of approaches in the CAL paradigm that achieve different performance/speedup trade-offs. Lastly, we would like to point out that some of the CAL methods are ablations of each other. For example, CAL-ER is ablation for CAL-DER (or CAL-SD) when we replace the distillation component. Similarly, CAL-SD is ablation of CAL-SDS2, where we remove the submodular selection part.\looseness-1

\subsection{Datasets and Experimental Setup}
\vspace{-.5\baselineskip}
We use the following datasets, which span a spectrum of data modalities, scale (both in terms of dataset size, and model's computational/memory footprint), and class balance.

{\bf FMNIST:} The FMNIST dataset consists of 70,000 28$\times$28 grayscale images of fashion items belonging to 10 classes~\citep{fmnist}. A ResNet-18 architecture~\citep{resnet} with SGD is used. We apply data augmentation, as in~\cite{distil2021}, consisting of random horizontal flips and random croppings.

{\bf CIFAR-10:} CIFAR-10 consists of 60,000 32$\times$32 color images with 10 different categories~\citep{cifar10}. We use a ResNet-18 and use the SGD optimizer for all CIFAR-10 experiments. We apply data augmentations consisting of random horizontal flips and random croppings.

{\bf MedMNIST:} We use the DermaMNIST dataset within the MedMNIST collection~\citep{medmnistv1, medmnistv2} for performance evaluation of CAL on medical imaging modalities. It consists of 3-color channel dermatoscopy images of 7 different skin diseases, originally obtained from ~\cite{https://doi.org/10.48550/arxiv.1902.03368, Tschandl2018TheHD}. A ResNet-18 architecture is used for all DermaMNIST experiments.

{\bf Amazon Polarity Review:} \citep{DBLP:journals/corr/ZhangZL15} is an NLP dataset consisting of reviews from Amazon and their corresponding star-ratings (5 classes) which was used for active learning in \citet{Coleman2020Selection}. Similar to the previous work we consider a total unlabeled pool of size 2 million sentences and use a VDCNN-9~\citep{Schwenk2017VeryDC} architecture trained using Adam.

{\bf COLA:} COLA \citep{warstadt2018neural} aims to check the linguistic acceptability of a sentence via binary classification. We consider an unlabeled size-7000 pool similar to~\citet{ein-dor-etal-2020-active} and use a BERT~\citep{Devlin2019BERTPO} backbone trained using Adam.

{\bf Single-Cell Cell Type Identity Classification:}
Recent single-cell RNA sequencing (scRNA-seq) technologies have enabled large-scale characterization of hundreds of thousands to millions of cells in complex tissues, and accurate cell type annotation is a crucial step in the study of such datasets. To this end, several deep learning models have been proposed to automatically label new scRNA-seq datasets~\citep{XIE20215874}. The HCL dataset is highly \emph{class-imbalanced} and consists of scRNA-seq data for 562,977 cells across 63 cell types represented in 56 human tissues~\citep{Han2020-sl}. We use the ACTINN model~\citep{10.1093/bioinformatics/btz592}, a four-layer multi-layer perceptron that predicts the cell-type for each cell given its expression of 28832 genes, and uses an SGD optimizer.

{\bf Hyperparameters:} Details about the specific hardware and the choices of hyperparameters used to train models for each technique can be found in Appendix \ref{sec:appen_hyperparameters}. 

{\bf Active Learning setup:} As done in previous work~\citep{Coleman2020Selection, glister2021} for CIFAR10 and Amazon polarity review, budgets go from 10\% to 50\% in increments of 10\% (results for other query sizes are presented in Appendix~\ref{sec:query_size_effect}). For FMNIST, MedMNIST, and Cell-type datasets, it goes from 10\% to 30\% in increments of 5\%. Lastly, for COLA, we follow a budget
using absolute sizes from 200 to 1000 in increments of 200 (similar to~\citet{ein-dor-etal-2020-active}).\footnote{We follow the query set sizes from~\citep{ein-dor-etal-2020-active}} We adopt the AL framework proposed in~\cite{distil2021} for all experiments. In the main paper, we here present results for an uncertainty sampling-based acquisition function. \emph{However, we provide results using other acquisition functions in Appendix~\ref{sec:appen_other_acquisition_func}}.
\vspace{-.5\baselineskip}

\subsection{Performance vs Speedup}
\label{sec:performance_vs_speedup}



\begin{figure}[tbh]
  \begin{minipage}[c]{0.6\textwidth}
    \includegraphics[width=\linewidth]{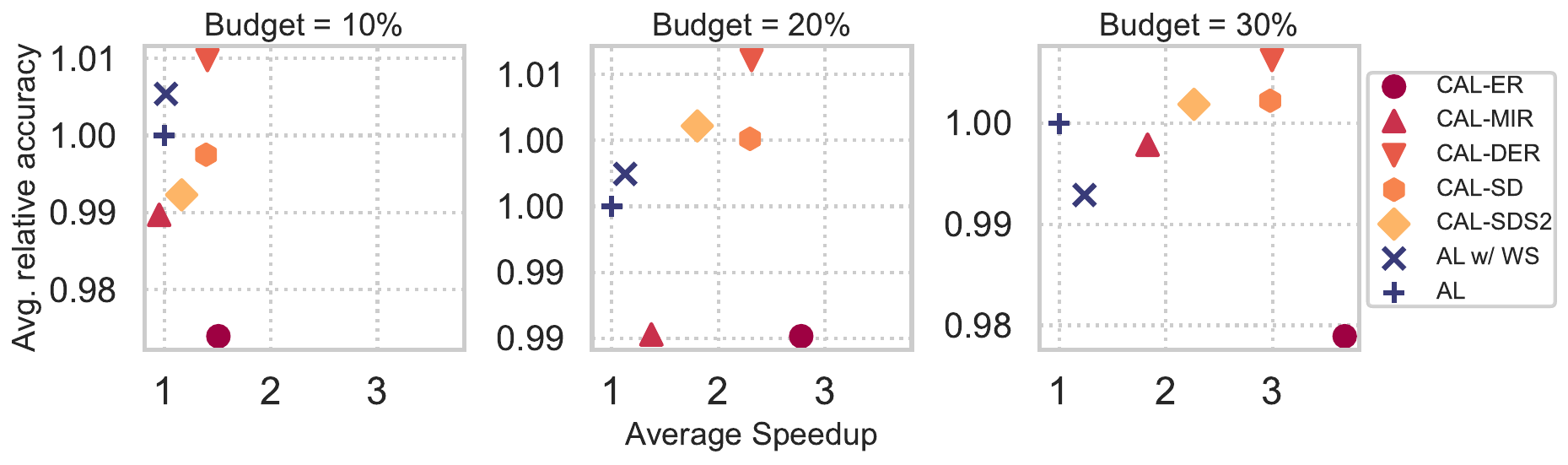}
  \end{minipage}\hfill
  \begin{minipage}[c]{0.35\textwidth}
    \caption{\small Relative accuracy vs.\ speedup averaged over different datasets. For every given budget, we have \emph{at least one} CAL method that is faster with the same (if not better) accuracy than the baseline.}
    \label{fig:all_budget}
  \end{minipage}
 \vspace{-.75\baselineskip}
\end{figure} 

Our goal is to show that for every dataset and budget, there exists at least one CAL method that performs equivalent (if not better) than baseline AL. However, with raw accuracies, it is difficult to compare different CAL methods and baseline AL over different datasets at different budgets. Therefore, we begin by observing the relative gain in accuracy over the AL baseline. Relative gains further make it feasible to take an average across the budgets, for a given dataset, and to take averages across the datasets, for a given budget. For \emph{every budget}, we normalize the accuracies of each method
by that of baseline AL. This makes the baseline accuracy always \textbf{1},
irrespective of budget and dataset. Relative performances greater than
1 indicate better than baseline accuracy (and the opposite for the less
than 1 case). Having said that: (1) keeping the budgets fixed to 10\%,
20\%, and 30\% and averaging over the datasets (except COLA, since it
has a different budget) give us Figure~\ref{fig:all_budget}; (2)
keeping the dataset fixed, averaging the relative accuracy vs.\
speedups across different budgets give us
Figure~\ref{fig:all_datasets}; and (3) further averaging the above
across different datasets give us Figure~\ref{fig:main_result}.
Methods in the top right corner are preferable.
For space reasons, we display only relative accuracies in the main paper,
but all absolute accuracies and standard deviations for each method for every dataset and every budget are available in Appendix~\ref{sec: Results in Tabular Form}.

From the overall and per-dataset depiction of CAL's performance (Figures~\ref{fig:main_result} and~\ref{fig:all_datasets}, respectively), it is evident that \emph{there exists a CAL method that attains a significant speedup over a standard AL technique for every dataset and budget while preserving test set accuracy.} From Figure~\ref{fig:all_datasets}, we can further see that for some datasets (such as FMNIST and CIFAR-10), CAL-ER, a non-distillation, and uniform sampling-based method, only incur a minor drop in performance but attain the highest speedup. This suggests that naively biasing learning towards recent tasks can be sufficient to adapt the model to a new set of points between AL rounds. However, as we show in Figure~\ref{fig:all_budget} it is not universally true for all the datasets (at different budgets). Hence, the methods which include some type of distillation term (CAL-DER, CAL-SD, CAL-SDS2) generally perform the best out of all CAL methods. We believe that the submodular sampling-based method (CAL-SDS2) can be accelerated using stochastic methods and results improved by considering other submodular functions, which we leave as future work. It should be mentioned, however, that
the concave function on $h(x_i)$ was essential for CAL-SDS2's performance.

\subsection{Comparison Between Standard and CAL Models}
\vspace{-.5\baselineskip}
In this section, we next assess whether CAL training has any adverse effect on the final model's behavior. We first demonstrate that CAL does \emph{not} result in any deterioration of model robustness (Section~\ref{Robustness}). We then demonstrate that CAL models and baseline trained models are uncertain about a similar set of unseen examples (Section~\ref{Correlation}). Lastly, in appendix~\ref{sec:appen_hyperparameters} we provide a sensitivity analysis of our proposed methods, where we demonstrate that \textit{CAL methods are robust to the changes to the hyperparameters.}

\subsubsection{Robustness}
\label{Robustness}
\vspace{-.5\baselineskip}
\begin{figure}
  \begin{minipage}[c]{0.45\textwidth}
    \includegraphics[width=\textwidth]{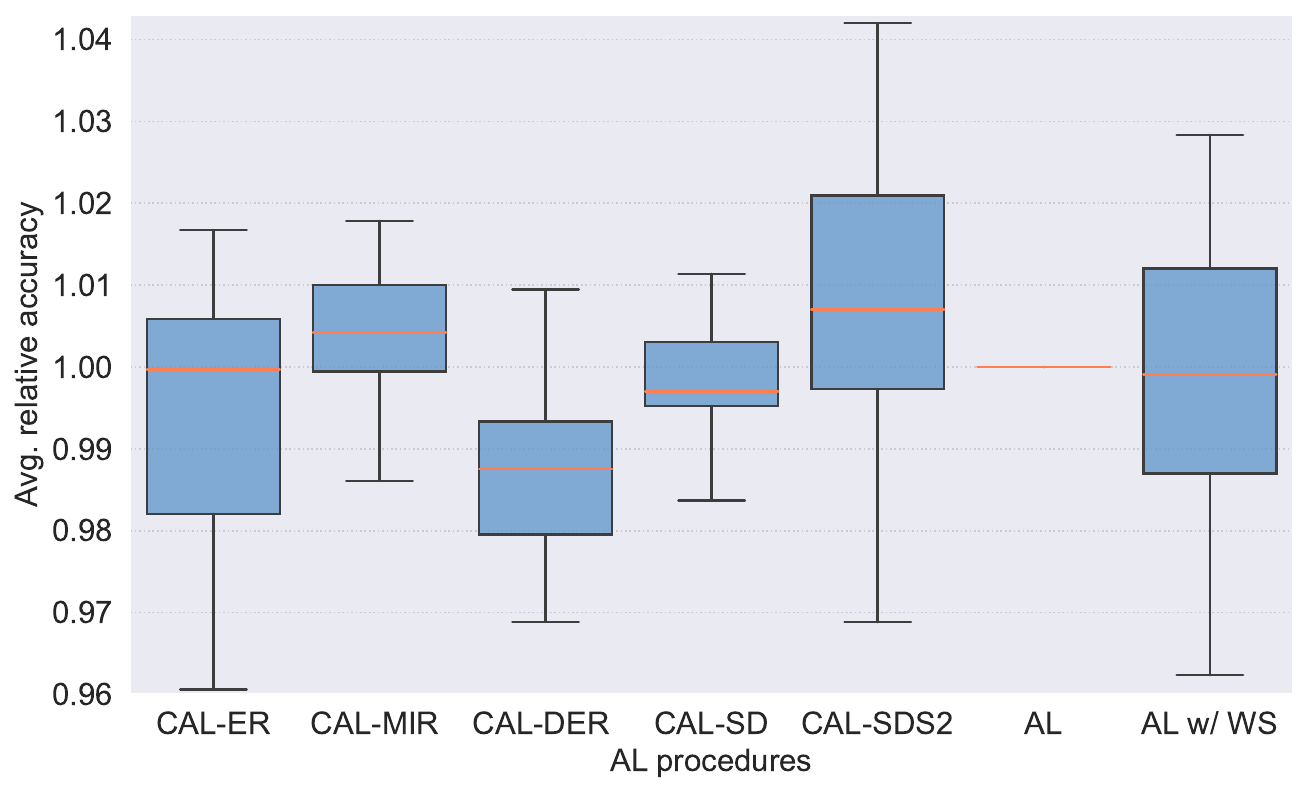}
  \end{minipage}\hfill
  \begin{minipage}[c]{0.5\textwidth}
    \caption{\small Comparison of CAL methods with the baseline on CIFAR-10C; absolute accuracies on individual benchmark and its average difference with the baseline AL are reported in Appendix~\ref{sec: Results in Tabular Form} table~\ref{tab: robustnesss} and \ref{tab:robustnesss_difference}, respectively. All CAL methods perform within a standard deviation of the standard AL baseline, and CAL-SDS2 achieves the highest robust accuracy on average.}
    \label{fig:robustness}
  \end{minipage}
 \vspace{-1\baselineskip}
\end{figure} 

Test time distributions can vary from training distributions, so it is important to ensure that models can generalize across different domains. Since models trained using CAL methods require significantly fewer gradient steps, the modified training procedure may produce fickler models that are less robust to domain shifts. To ensure against this, we evaluate CAL-method-trained model robustness in this section. We consider CIFAR-10C~\cite{hendrycks2019robustness}, a dataset comprising 19 different corruptions each done at 5 levels of severity. For each model trained up to a $50\%$ budget, we report the average classification accuracy over each corruption and compare it against the baseline in Figure~\ref{fig:robustness}; each result is an average of over three random seeds. We note that most of the CAL methods perform statistically similarly to standard active learning, all while providing significant acceleration. Moreover, on average across all the tests, {\bf models trained with CAL-SDS2 are better than the models trained with CAL-DER}, where we see a difference of as much as 5\% in corruptions such as \emph{glass blur}; please refer to appendix~\ref{sec:appen_OOD} table~\ref{tab: robustnesss}, table~\ref{tab:robustness_significance} and \ref{tab:robustnesss_difference} for the absolute accuracies and statistical tests. Submodular sampling replays a diverse representative subset of history which is likely the reason behind CAL-SDS2's better robustness. The relationship of diversity with robustness has also been explored in previous works including \cite{DBLP:journals/corr/abs-2106-07760, fang2022data, rozen-etal-2019-diversify, DBLP:journals/corr/abs-1807-01477}.


\subsubsection{Correlation of Uncertainty Scores}
\label{Correlation}
\vspace{-.5\baselineskip}

\looseness-1

For models trained using CAL techniques to be used as valid substitutes for standard AL models, these two classes of models need to query similar samples at each AL round. This is particularly important if AL is being used solely as a data subset selection procedure (where the user is concerned about the quality of the resulting labeled dataset as opposed to the final model). When using uncertainty sampling as the AL acquisition function, computing the Pearson correlation between the entropy scores of baseline and CAL models on the validation set after every query round is one way of determining this. Since most AL policies incorporate some notion of uncertainty, we hypothesize that the results of this experiment should extend to other acquisition functions as well. A similar analysis is done in~\cite{Coleman2020Selection}.

\begin{wrapfigure}[18]{r}{0.5\textwidth}
    \vskip -0.15in	
    \centering	
    \includegraphics[width=0.5\textwidth]{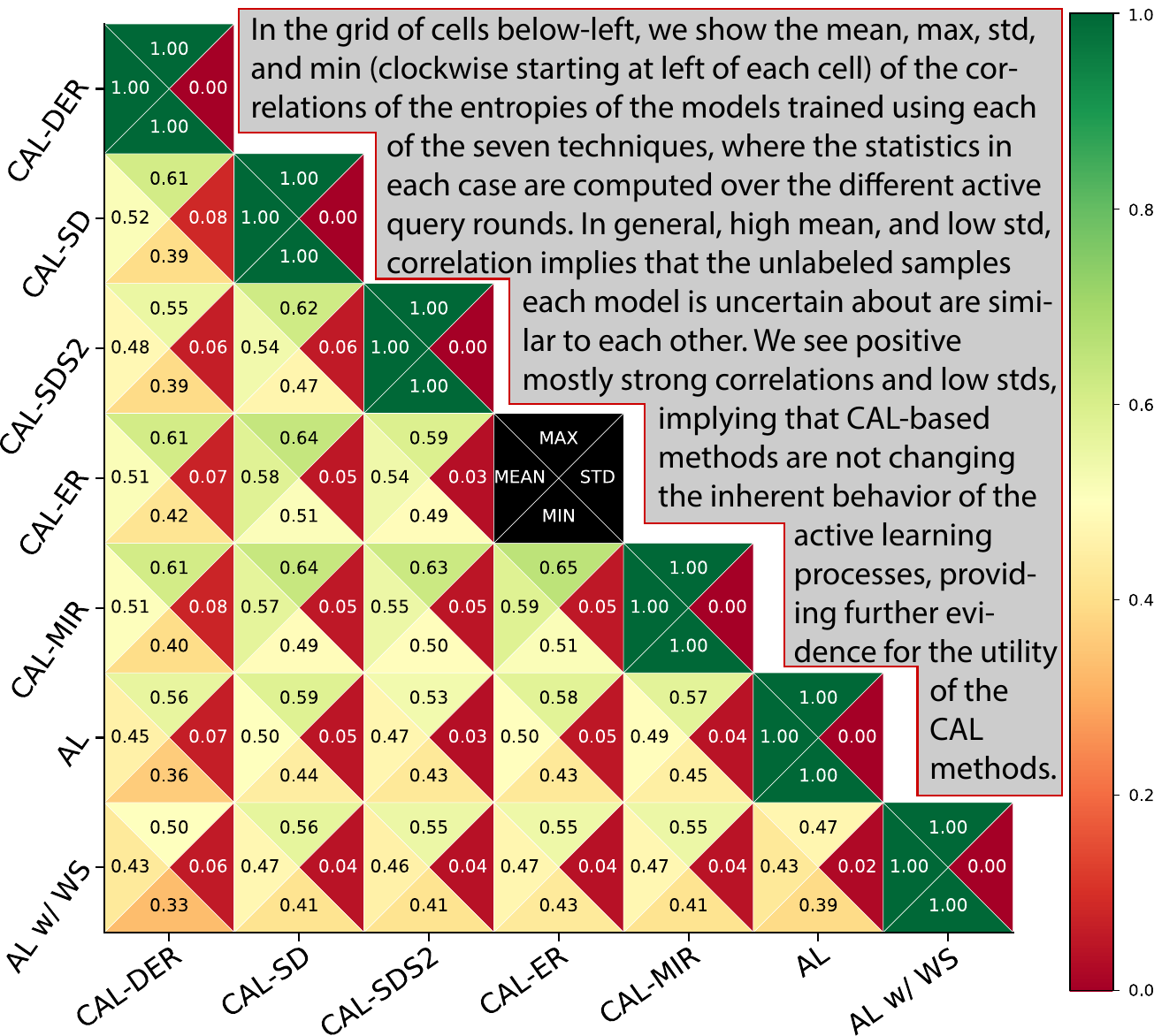}
    \caption{\small Cross-Method Entropy Correlation Statistics.}
    \label{fig:score_corr}
\end{wrapfigure}

As seen in Figure~\ref{fig:score_corr}, the Pearson correlation between all pairs of models is positive at every AL query round. Thus, the nature of samples that the models are uncertain about, and thus are likely to be chosen at each round, is similar between the CAL-trained models and baseline AL-trained models. A breakdown of these correlations at every round is provided in Figure~\ref{fig:all_correlation} in the Appendix.

\vspace{-.9\baselineskip}
\section{Conclusion and Future Work}
\vspace{-.5\baselineskip}
We proposed the CAL framework, the first method to circumvent the
problem of having to retrain models between batch AL rounds. Across
vision, natural language, medical imaging, and biological datasets, we
show there is always a CAL-based method that either matches or
outperforms standard AL while achieving considerable speedups. Since
CAL is independent of the model architecture and AL strategy, this
framework applies to a broad range of settings.  

\paragraph{Empirical Extensions}
Future empirical directions may include the following: {\bf (1):} CAL reduces the training
time of the model, but not the AL query time although query time
reductions could be an offshoot of this work; {\bf (2):} CAL operates using
existing AL query acquisition functions, but it is possible to
tailor acquisition functions for CAL methods yielding further generalization and computational
improvements; {\bf (3):} while
SDS2's use of submodularity helped robustness, additional submodular
strategies can be used to further improve results and also for diverse
AL query selection; and {\bf (4):} CAL provides a novel application for CL;
future CL work can be partially assessed based on its CAL
performance.

\paragraph{Theoretical Extensions}

The discrepancy-based Active Learning framework is promising to establish generalization error bounds for the replay-based CAL framework. Specifically, leveraging the generalization results presented in Theorem~2 of \cite{discre}, the upper bound on the risk function based on true data distribution (referred to as $\mathbb{P}$ in the theorem) is contingent upon the empirical distribution of the presently labeled data (referred to as $\mathbb{Q}$ in the theorem), that includes the newly acquired query batch.

In contrast to utilizing the entirety of the available labeled data, CAL strategically opts for a subset of this data through the employment of replay methods, for acceleration. Due to this, the empirical distribution of examples chosen by CAL (call it $\mathbb{Q}^\prime $) could deviate from $\mathbb{Q}$. Consequently, one can modify the upper bound of the risk function based on true data distribution in Theorem~2 by adding the error term (which would depend on some notion of distance between $\mathbb{Q}$ and $\mathbb{Q}^\prime$) which vanishes when we consider all the currently labeled examples, thus, recovering standard AL.

\subsubsection*{Acknowledgments}
This work was supported in part by the CONIX Research Center, one of six centers in JUMP, a
Semiconductor Research Corporation (SRC) program sponsored by DARPA, by the National Science Foundation under Grant Nos.\ IIS-2106937 and IIS-2148367, and by NIH/NHGRI U01 award HG009395. 
We thank Lilly Kumari, Tianyi Zhou, Shengjie Wang and all other MELODI lab members for their helpful discussions and feedback. We also thank the TMLR Action Editor and reviewers for their constructive comments.

\bibliography{main}

\begin{thebibliography}{96}
\providecommand{\natexlab}[1]{#1}
\providecommand{\url}[1]{\texttt{#1}}
\expandafter\ifx\csname urlstyle\endcsname\relax
  \providecommand{\doi}[1]{doi: #1}\else
  \providecommand{\doi}{doi: \begingroup \urlstyle{rm}\Url}\fi

\bibitem[Abraham \& Robins(2005)Abraham and Robins]{Abraham2005}
Wickliffe~C Abraham and Anthony Robins.
\newblock Memory retention--the synaptic stability versus plasticity dilemma.
\newblock \emph{Trends Neurosci}, 28\penalty0 (2):\penalty0 73--78, February 2005.

\bibitem[Aljundi et~al.(2019{\natexlab{a}})Aljundi, Caccia, Belilovsky, Caccia, Lin, Charlin, and Tuytelaars]{mir2019}
Rahaf Aljundi, Lucas Caccia, Eugene Belilovsky, Massimo Caccia, Min Lin, Laurent Charlin, and Tinne Tuytelaars.
\newblock Online continual learning with maximally interfered retrieval.
\newblock \emph{CoRR}, abs/1908.04742, 2019{\natexlab{a}}.
\newblock URL \url{http://arxiv.org/abs/1908.04742}.

\bibitem[Aljundi et~al.(2019{\natexlab{b}})Aljundi, Lin, Goujaud, and Bengio]{aljundi2019gradient}
Rahaf Aljundi, Min Lin, Baptiste Goujaud, and Yoshua Bengio.
\newblock Gradient based sample selection for online continual learning.
\newblock \emph{arXiv preprint arXiv:1903.08671}, 2019{\natexlab{b}}.

\bibitem[Ash \& Adams(2020)Ash and Adams]{Ash2020}
Jordan Ash and Ryan~P Adams.
\newblock On warm-starting neural network training.
\newblock In H.~Larochelle, M.~Ranzato, R.~Hadsell, M.F. Balcan, and H.~Lin (eds.), \emph{Advances in Neural Information Processing Systems}, volume~33, pp.\  3884--3894. Curran Associates, Inc., 2020.
\newblock URL \url{https://proceedings.neurips.cc/paper/2020/file/288cd2567953f06e460a33951f55daaf-Paper.pdf}.

\bibitem[Ash et~al.(2020)Ash, Zhang, Krishnamurthy, Langford, and Agarwal]{badge2020}
Jordan~T. Ash, Chicheng Zhang, Akshay Krishnamurthy, John Langford, and Alekh Agarwal.
\newblock Deep batch active learning by diverse, uncertain gradient lower bounds.
\newblock \emph{ArXiv}, abs/1906.03671, 2020.

\bibitem[Atlas et~al.(1989)Atlas, Cohn, and Ladner]{atlas1989training}
Les Atlas, David Cohn, and Richard Ladner.
\newblock Training connectionist networks with queries and selective sampling.
\newblock \emph{Advances in neural information processing systems}, 2, 1989.

\bibitem[Ayub \& Fendley(2022)Ayub and Fendley]{FoCAL2022}
Ali Ayub and Carter Fendley.
\newblock Few-shot continual active learning by a robot, 2022.
\newblock URL \url{https://arxiv.org/abs/2210.04137}.

\bibitem[Balcan et~al.(2010)Balcan, Hanneke, and Vaughan]{Balcan2010}
Maria-Florina Balcan, Steve Hanneke, and Jennifer~Wortman Vaughan.
\newblock The true sample complexity of active learning.
\newblock \emph{Machine Learning}, 80\penalty0 (2-3):\penalty0 111--139, April 2010.
\newblock \doi{10.1007/s10994-010-5174-y}.
\newblock URL \url{https://doi.org/10.1007/s10994-010-5174-y}.

\bibitem[Beck et~al.(2021)Beck, Sivasubramanian, Dani, Ramakrishnan, and Iyer]{distil2021}
Nathan Beck, Durga Sivasubramanian, Apurva Dani, Ganesh Ramakrishnan, and Rishabh~K. Iyer.
\newblock Effective evaluation of deep active learning on image classification tasks.
\newblock \emph{CoRR}, abs/2106.15324, 2021.
\newblock URL \url{https://arxiv.org/abs/2106.15324}.

\bibitem[Bender et~al.(2021)Bender, Gebru, McMillan-Major, and Shmitchell]{stochasticparrot2021}
Emily~M. Bender, Timnit Gebru, Angelina McMillan-Major, and Shmargaret Shmitchell.
\newblock On the dangers of stochastic parrots: Can language models be too big?
\newblock In \emph{Proceedings of the 2021 ACM Conference on Fairness, Accountability, and Transparency}, FAccT '21, pp.\  610–623, New York, NY, USA, 2021. Association for Computing Machinery.
\newblock ISBN 9781450383097.
\newblock \doi{10.1145/3442188.3445922}.
\newblock URL \url{https://doi.org/10.1145/3442188.3445922}.

\bibitem[Bilmes(2022)]{submodular2022}
Jeff~A. Bilmes.
\newblock Submodularity in machine learning and artificial intelligence.
\newblock \emph{CoRR}, abs/2202.00132, 2022.
\newblock URL \url{https://arxiv.org/abs/2202.00132}.

\bibitem[Buzzega et~al.(2020)Buzzega, Boschini, Porrello, Abati, and CALDERARA]{DER2020}
Pietro Buzzega, Matteo Boschini, Angelo Porrello, Davide Abati, and SIMONE CALDERARA.
\newblock Dark experience for general continual learning: a strong, simple baseline.
\newblock In H.~Larochelle, M.~Ranzato, R.~Hadsell, M.~F. Balcan, and H.~Lin (eds.), \emph{Advances in Neural Information Processing Systems}, volume~33, pp.\  15920--15930. Curran Associates, Inc., 2020.

\bibitem[Chaudhry et~al.(2019)Chaudhry, Ranzato, Rohrbach, and Elhoseiny]{AGEM2019}
Arslan Chaudhry, Marc\textquotesingle~Aurelio Ranzato, Marcus Rohrbach, and Mohamed Elhoseiny.
\newblock Efficient lifelong learning with a-gem.
\newblock In \emph{ICLR}, 2019.

\bibitem[Chaudhry et~al.(2020)Chaudhry, Gordo, Lopez-Paz, Dokania, and Torr]{chaudhry2020using}
Arslan Chaudhry, Albert Gordo, David Lopez-Paz, Puneet~K. Dokania, and Philip Torr.
\newblock Using hindsight to anchor past knowledge in continual learning, 2020.
\newblock URL \url{https://openreview.net/forum?id=Hke12T4KPS}.

\bibitem[Codella et~al.(2019)Codella, Rotemberg, Tschandl, Celebi, Dusza, Gutman, Helba, Kalloo, Liopyris, Marchetti, Kittler, and Halpern]{https://doi.org/10.48550/arxiv.1902.03368}
Noel Codella, Veronica Rotemberg, Philipp Tschandl, M.~Emre Celebi, Stephen Dusza, David Gutman, Brian Helba, Aadi Kalloo, Konstantinos Liopyris, Michael Marchetti, Harald Kittler, and Allan Halpern.
\newblock Skin lesion analysis toward melanoma detection 2018: A challenge hosted by the international skin imaging collaboration (isic), 2019.
\newblock URL \url{https://arxiv.org/abs/1902.03368}.

\bibitem[Cohn et~al.(1994)Cohn, Atlas, and Ladner]{CohnAL94}
David~A. Cohn, Les~E. Atlas, and Richard~E. Ladner.
\newblock Improving generalization with active learning.
\newblock \emph{Mach. Learn.}, 15\penalty0 (2):\penalty0 201--221, 1994.
\newblock URL \url{http://dblp.uni-trier.de/db/journals/ml/ml15.html#CohnAL94}.

\bibitem[Coleman et~al.(2020{\natexlab{a}})Coleman, Chou, Culatana, Bailis, Berg, Sumbaly, Zaharia, and Yalniz]{DBLP:journals/corr/abs-2007-00077}
Cody Coleman, Edward Chou, Sean Culatana, Peter Bailis, Alexander~C. Berg, Roshan Sumbaly, Matei Zaharia, and I.~Zeki Yalniz.
\newblock Similarity search for efficient active learning and search of rare concepts.
\newblock \emph{CoRR}, abs/2007.00077, 2020{\natexlab{a}}.
\newblock URL \url{https://arxiv.org/abs/2007.00077}.

\bibitem[Coleman et~al.(2020{\natexlab{b}})Coleman, Yeh, Mussmann, Mirzasoleiman, Bailis, Liang, Leskovec, and Zaharia]{Coleman2020Selection}
Cody Coleman, Christopher Yeh, Stephen Mussmann, Baharan Mirzasoleiman, Peter Bailis, Percy Liang, Jure Leskovec, and Matei Zaharia.
\newblock Selection via proxy: Efficient data selection for deep learning.
\newblock In \emph{International Conference on Learning Representations}, 2020{\natexlab{b}}.
\newblock URL \url{https://openreview.net/forum?id=HJg2b0VYDr}.

\bibitem[Cui \& Sato(2020)Cui and Sato]{discre}
Zhenghang Cui and Issei Sato.
\newblock Active learning using discrepancy.
\newblock \url{https://realworldml.github.io/files/cr/7_cui_paper.pdf}, 2020.
\newblock [Accessed 21-11-2023].

\bibitem[Dagan \& Engelson(1995)Dagan and Engelson]{dagan1995}
Ido Dagan and Sean~P. Engelson.
\newblock Committee-based sampling for training probabilistic classifiers.
\newblock In \emph{Proceedings of the Twelfth International Conference on International Conference on Machine Learning}, ICML'95, pp.\  150–157, San Francisco, CA, USA, 1995. Morgan Kaufmann Publishers Inc.
\newblock ISBN 1558603778.

\bibitem[De~Lange et~al.(2022)De~Lange, Aljundi, Masana, Parisot, Jia, Leonardis, Slabaugh, and Tuytelaars]{CLSurvey2022}
Matthias De~Lange, Rahaf Aljundi, Marc Masana, Sarah Parisot, Xu~Jia, Aleš Leonardis, Gregory Slabaugh, and Tinne Tuytelaars.
\newblock A continual learning survey: Defying forgetting in classification tasks.
\newblock \emph{IEEE Transactions on Pattern Analysis and Machine Intelligence}, 44\penalty0 (7):\penalty0 3366--3385, 2022.
\newblock \doi{10.1109/TPAMI.2021.3057446}.

\bibitem[Devlin et~al.(2019)Devlin, Chang, Lee, and Toutanova]{Devlin2019BERTPO}
Jacob Devlin, Ming-Wei Chang, Kenton Lee, and Kristina Toutanova.
\newblock Bert: Pre-training of deep bidirectional transformers for language understanding.
\newblock \emph{ArXiv}, abs/1810.04805, 2019.

\bibitem[Dhar(2020)]{Dhar2020}
Payal Dhar.
\newblock The carbon impact of artificial intelligence.
\newblock \emph{Nature Machine Intelligence}, 2\penalty0 (8):\penalty0 423--425, Aug 2020.
\newblock ISSN 2522-5839.
\newblock \doi{10.1038/s42256-020-0219-9}.
\newblock URL \url{https://doi.org/10.1038/s42256-020-0219-9}.

\bibitem[Ein-Dor et~al.(2020)Ein-Dor, Halfon, Gera, Shnarch, Dankin, Choshen, Danilevsky, Aharonov, Katz, and Slonim]{ein-dor-etal-2020-active}
Liat Ein-Dor, Alon Halfon, Ariel Gera, Eyal Shnarch, Lena Dankin, Leshem Choshen, Marina Danilevsky, Ranit Aharonov, Yoav Katz, and Noam Slonim.
\newblock {A}ctive {L}earning for {BERT}: {A}n {E}mpirical {S}tudy.
\newblock In \emph{Proceedings of the 2020 Conference on Empirical Methods in Natural Language Processing (EMNLP)}, pp.\  7949--7962, Online, November 2020. Association for Computational Linguistics.
\newblock \doi{10.18653/v1/2020.emnlp-main.638}.
\newblock URL \url{https://aclanthology.org/2020.emnlp-main.638}.

\bibitem[Ertekin et~al.(2007)Ertekin, Huang, Bottou, and Giles]{10.1145/1321440.1321461}
Seyda Ertekin, Jian Huang, Leon Bottou, and Lee Giles.
\newblock Learning on the border: Active learning in imbalanced data classification.
\newblock In \emph{Proceedings of the Sixteenth ACM Conference on Conference on Information and Knowledge Management}, CIKM '07, pp.\  127--136, New York, NY, USA, 2007. Association for Computing Machinery.
\newblock ISBN 9781595938039.
\newblock \doi{10.1145/1321440.1321461}.
\newblock URL \url{https://doi.org/10.1145/1321440.1321461}.

\bibitem[Fang et~al.(2022)Fang, Ilharco, Wortsman, Wan, Shankar, Dave, and Schmidt]{fang2022data}
Alex Fang, Gabriel Ilharco, Mitchell Wortsman, Yuhao Wan, Vaishaal Shankar, Achal Dave, and Ludwig Schmidt.
\newblock Data determines distributional robustness in contrastive language image pre-training (clip), 2022.

\bibitem[Farquhar \& Gal(2018)Farquhar and Gal]{farquhar2018towards}
Sebastian Farquhar and Yarin Gal.
\newblock Towards robust evaluations of continual learning.
\newblock \emph{arXiv preprint arXiv:1805.09733}, 2018.

\bibitem[Fisher et~al.(1978)Fisher, Nemhauser, and Wolsey]{fisher1978analysis}
M.L. Fisher, G.L. Nemhauser, and L.A. Wolsey.
\newblock An analysis of approximations for maximizing submodular set functions---{II}.
\newblock In \emph{Polyhedral combinatorics}, 1978.

\bibitem[French(1999)]{french1999catastrophic}
Robert~M French.
\newblock Catastrophic forgetting in connectionist networks.
\newblock \emph{Trends in cognitive sciences}, 3\penalty0 (4):\penalty0 128--135, 1999.

\bibitem[Gong et~al.(2018)Gong, Zhong, and Hu]{DBLP:journals/corr/abs-1807-01477}
Zhiqiang Gong, Ping Zhong, and Weidong Hu.
\newblock Diversity in machine learning.
\newblock \emph{CoRR}, abs/1807.01477, 2018.
\newblock URL \url{http://arxiv.org/abs/1807.01477}.

\bibitem[Guillory et~al.(2009)Guillory, Chastain, and Bilmes]{guillory2009-actnoncon}
Andrew Guillory, Erick Chastain, and Jeff Bilmes.
\newblock Active learning as non-convex optimization.
\newblock In \emph{Twelfth International Conference on Artificial Intelligence and Statistics (AISTAT)}, Clearwater Beach, Florida, April 2009.

\bibitem[Han et~al.(2020)Han, Zhou, Fei, Sun, Wang, Chen, Chen, Wang, Tang, Ge, Zhou, Ye, Jiang, Wu, Xiao, Jia, Zhang, Ma, Zhang, Bai, Lai, Yu, Zhu, Lin, Gao, Wang, Wu, Zhang, Zhan, Zhu, Hu, Wang, Chen, Huang, Liang, Chen, Wang, Zhang, and Guo]{Han2020-sl}
Xiaoping Han, Ziming Zhou, Lijiang Fei, Huiyu Sun, Renying Wang, Yao Chen, Haide Chen, Jingjing Wang, Huanna Tang, Wenhao Ge, Yincong Zhou, Fang Ye, Mengmeng Jiang, Junqing Wu, Yanyu Xiao, Xiaoning Jia, Tingyue Zhang, Xiaojie Ma, Qi~Zhang, Xueli Bai, Shujing Lai, Chengxuan Yu, Lijun Zhu, Rui Lin, Yuchi Gao, Min Wang, Yiqing Wu, Jianming Zhang, Renya Zhan, Saiyong Zhu, Hailan Hu, Changchun Wang, Ming Chen, He~Huang, Tingbo Liang, Jianghua Chen, Weilin Wang, Dan Zhang, and Guoji Guo.
\newblock Construction of a human cell landscape at single-cell level.
\newblock \emph{Nature}, 581\penalty0 (7808):\penalty0 303--309, March 2020.

\bibitem[Hanneke(2007)]{10.1145/1273496.1273541}
Steve Hanneke.
\newblock A bound on the label complexity of agnostic active learning.
\newblock In \emph{Proceedings of the 24th International Conference on Machine Learning}, ICML '07, pp.\  353–360, New York, NY, USA, 2007. Association for Computing Machinery.
\newblock ISBN 9781595937933.
\newblock \doi{10.1145/1273496.1273541}.
\newblock URL \url{https://doi.org/10.1145/1273496.1273541}.

\bibitem[Hanneke(2009)]{hanneke2009theoretical}
Steve Hanneke.
\newblock \emph{Theoretical foundations of active learning}.
\newblock Carnegie Mellon University, 2009.

\bibitem[Hanneke(2012)]{JMLR:v13:hanneke12a}
Steve Hanneke.
\newblock Activized learning: Transforming passive to active with improved label complexity.
\newblock \emph{Journal of Machine Learning Research}, 13\penalty0 (49):\penalty0 1469--1587, 2012.
\newblock URL \url{http://jmlr.org/papers/v13/hanneke12a.html}.

\bibitem[He et~al.(2016)He, Zhang, Ren, and Sun]{resnet}
Kaiming He, Xiangyu Zhang, Shaoqing Ren, and Jian Sun.
\newblock Deep residual learning for image recognition.
\newblock In \emph{2016 IEEE Conference on Computer Vision and Pattern Recognition (CVPR)}, pp.\  770--778, 2016.
\newblock \doi{10.1109/CVPR.2016.90}.

\bibitem[Hendrycks \& Dietterich(2019)Hendrycks and Dietterich]{hendrycks2019robustness}
Dan Hendrycks and Thomas Dietterich.
\newblock Benchmarking neural network robustness to common corruptions and perturbations.
\newblock \emph{Proceedings of the International Conference on Learning Representations}, 2019.

\bibitem[Jia et~al.(2022)Jia, Unger, Wu, Lin, Baines, Efrain, Prajapati, McCormick, Mohd-Yusof, Luo, Mudigere, Park, Smelyanskiy, and Aiken.]{unity2022}
Zhihao Jia, Colin Unger, Wei Wu, Sina Lin, Mandeep Baines, Vinay Ramakrishnaiah~Carlos Efrain, Nirmal Prajapati, Pat McCormick, Jamaludin Mohd-Yusof, Xi~Luo, Dheevatsa Mudigere, Jongsoo Park, Misha Smelyanskiy, and Alex Aiken.
\newblock Unity: Accelerating {DNN} training through joint optimization of algebraic transformations and parallelization.
\newblock In \emph{16th USENIX Symposium on Operating Systems Design and Implementation (OSDI 22)}, Carlsbad, CA, July 2022. USENIX Association.
\newblock URL \url{https://www.usenix.org/conference/osdi22/presentation/jia}.

\bibitem[Killamsetty et~al.(2021{\natexlab{a}})Killamsetty, Sivasubramanian, Ramakrishnan, and Iyer]{glister2021}
KrishnaTeja Killamsetty, Durga Sivasubramanian, Ganesh Ramakrishnan, and Rishabh~K. Iyer.
\newblock {GLISTER:} generalization based data subset selection for efficient and robust learning.
\newblock In \emph{Thirty-Fifth {AAAI} Conference on Artificial Intelligence, {AAAI} 2021, Thirty-Third Conference on Innovative Applications of Artificial Intelligence, {IAAI} 2021, The Eleventh Symposium on Educational Advances in Artificial Intelligence, {EAAI} 2021, Virtual Event, February 2-9, 2021}, pp.\  8110--8118. {AAAI} Press, 2021{\natexlab{a}}.
\newblock URL \url{https://ojs.aaai.org/index.php/AAAI/article/view/16988}.

\bibitem[Killamsetty et~al.(2021{\natexlab{b}})Killamsetty, Zhao, Chen, and Iyer]{DBLP:journals/corr/abs-2106-07760}
KrishnaTeja Killamsetty, Xujiang Zhao, Feng Chen, and Rishabh~K. Iyer.
\newblock {RETRIEVE:} coreset selection for efficient and robust semi-supervised learning.
\newblock \emph{CoRR}, abs/2106.07760, 2021{\natexlab{b}}.
\newblock URL \url{https://arxiv.org/abs/2106.07760}.

\bibitem[Kirkpatrick et~al.(2017{\natexlab{a}})Kirkpatrick, Pascanu, Rabinowitz, Veness, Desjardins, Rusu, Milan, Quan, Ramalho, Grabska-Barwinska, Hassabis, Clopath, Kumaran, and Hadsell]{doi:10.1073/pnas.1611835114}
James Kirkpatrick, Razvan Pascanu, Neil Rabinowitz, Joel Veness, Guillaume Desjardins, Andrei~A. Rusu, Kieran Milan, John Quan, Tiago Ramalho, Agnieszka Grabska-Barwinska, Demis Hassabis, Claudia Clopath, Dharshan Kumaran, and Raia Hadsell.
\newblock Overcoming catastrophic forgetting in neural networks.
\newblock \emph{Proceedings of the National Academy of Sciences}, 114\penalty0 (13):\penalty0 3521--3526, 2017{\natexlab{a}}.
\newblock \doi{10.1073/pnas.1611835114}.
\newblock URL \url{https://www.pnas.org/doi/abs/10.1073/pnas.1611835114}.

\bibitem[Kirkpatrick et~al.(2017{\natexlab{b}})Kirkpatrick, Pascanu, Rabinowitz, Veness, Desjardins, Rusu, Milan, Quan, Ramalho, Grabska-Barwinska, Hassabis, Clopath, Kumaran, and Hadsell]{ewc2017}
James Kirkpatrick, Razvan Pascanu, Neil Rabinowitz, Joel Veness, Guillaume Desjardins, Andrei~A. Rusu, Kieran Milan, John Quan, Tiago Ramalho, Agnieszka Grabska-Barwinska, Demis Hassabis, Claudia Clopath, Dharshan Kumaran, and Raia Hadsell.
\newblock Overcoming catastrophic forgetting in neural networks.
\newblock \emph{Proceedings of the National Academy of Sciences}, 114\penalty0 (13):\penalty0 3521--3526, 2017{\natexlab{b}}.
\newblock \doi{10.1073/pnas.1611835114}.

\bibitem[Kirkpatrick et~al.(2017{\natexlab{c}})Kirkpatrick, Pascanu, Rabinowitz, Veness, Desjardins, Rusu, Milan, Quan, Ramalho, Grabska-Barwinska, et~al.]{kirkpatrick2017overcoming}
James Kirkpatrick, Razvan Pascanu, Neil Rabinowitz, Joel Veness, Guillaume Desjardins, Andrei~A Rusu, Kieran Milan, John Quan, Tiago Ramalho, Agnieszka Grabska-Barwinska, et~al.
\newblock Overcoming catastrophic forgetting in neural networks.
\newblock \emph{Proceedings of the national academy of sciences}, 114\penalty0 (13):\penalty0 3521--3526, 2017{\natexlab{c}}.

\bibitem[Kirsch et~al.(2019)Kirsch, van Amersfoort, and Gal]{NEURIPS2019_95323660}
Andreas Kirsch, Joost van Amersfoort, and Yarin Gal.
\newblock Batchbald: Efficient and diverse batch acquisition for deep bayesian active learning.
\newblock In H.~Wallach, H.~Larochelle, A.~Beygelzimer, F.~d\textquotesingle Alch\'{e}-Buc, E.~Fox, and R.~Garnett (eds.), \emph{Advances in Neural Information Processing Systems}, volume~32. Curran Associates, Inc., 2019.
\newblock URL \url{https://proceedings.neurips.cc/paper/2019/file/95323660ed2124450caaac2c46b5ed90-Paper.pdf}.

\bibitem[Krizhevsky(2009)]{cifar10}
Alex Krizhevsky.
\newblock Learning multiple layers of features from tiny images.
\newblock pp.\  32--33, 2009.
\newblock URL \url{https://www.cs.toronto.edu/~kriz/learning-features-2009-TR.pdf}.

\bibitem[Lewis \& Catlett(1994)Lewis and Catlett]{LEWIS1994148}
David~D. Lewis and Jason Catlett.
\newblock Heterogeneous uncertainty sampling for supervised learning.
\newblock In William~W. Cohen and Haym Hirsh (eds.), \emph{Machine Learning Proceedings 1994}, pp.\  148--156. Morgan Kaufmann, San Francisco (CA), 1994.
\newblock ISBN 978-1-55860-335-6.
\newblock \doi{https://doi.org/10.1016/B978-1-55860-335-6.50026-X}.
\newblock URL \url{https://www.sciencedirect.com/science/article/pii/B978155860335650026X}.

\bibitem[Lewis \& Gale(1994)Lewis and Gale]{lewis1994}
David~D. Lewis and William~A. Gale.
\newblock A sequential algorithm for training text classifiers, 1994.
\newblock URL \url{http://arxiv.org/abs/cmp-lg/9407020}.
\newblock active learning roots.

\bibitem[Li et~al.(2021)Li, Krishnan, Wu, Kolouri, Pilly, and Braverman]{DBLP:journals/corr/abs-2104-08604}
Haoran Li, Aditya Krishnan, Jingfeng Wu, Soheil Kolouri, Praveen~K. Pilly, and Vladimir Braverman.
\newblock Lifelong learning with sketched structural regularization.
\newblock \emph{CoRR}, abs/2104.08604, 2021.
\newblock URL \url{https://arxiv.org/abs/2104.08604}.

\bibitem[Li et~al.(2020)Li, Sahu, Talwalkar, and Smith]{li2020federated}
Tian Li, Anit~Kumar Sahu, Ameet Talwalkar, and Virginia Smith.
\newblock Federated learning: Challenges, methods, and future directions.
\newblock \emph{{IEEE} signal processing magazine}, 37\penalty0 (3):\penalty0 50--60, 2020.

\bibitem[Li \& Bilmes(2007)Li and Bilmes]{li2007bayesian}
Xiao Li and Jeff Bilmes.
\newblock A {B}ayesian divergence prior for classifier adaptation.
\newblock In \emph{Artificial Intelligence and Statistics}, pp.\  275--282. PMLR, 2007.

\bibitem[Li \& Hoiem(2017)Li and Hoiem]{lwf2017}
Zhizhong Li and Derek Hoiem.
\newblock Learning without forgetting, 2017.

\bibitem[Lin \& Bilmes(2011)Lin and Bilmes]{lin-bilmes-2011-class}
Hui Lin and Jeff Bilmes.
\newblock A class of submodular functions for document summarization.
\newblock In \emph{Proceedings of the 49th Annual Meeting of the Association for Computational Linguistics: Human Language Technologies}, pp.\  510--520, Portland, Oregon, USA, June 2011. Association for Computational Linguistics.
\newblock URL \url{https://aclanthology.org/P11-1052}.

\bibitem[Liu et~al.(2013)Liu, Wei, Kirchhoff, Song, and Bilmes]{featurebased2013}
Yuzong Liu, Kai Wei, Katrin Kirchhoff, Yisong Song, and Jeff Bilmes.
\newblock Submodular feature selection for high-dimensional acoustic score spaces.
\newblock In \emph{2013 IEEE International Conference on Acoustics, Speech and Signal Processing}, pp.\  7184--7188, 2013.
\newblock \doi{10.1109/ICASSP.2013.6639057}.

\bibitem[Lopez-Paz \& Ranzato(2017)Lopez-Paz and Ranzato]{GEM2017}
David Lopez-Paz and Marc\textquotesingle~Aurelio Ranzato.
\newblock Gradient episodic memory for continual learning.
\newblock In I.~Guyon, U.~V. Luxburg, S.~Bengio, H.~Wallach, R.~Fergus, S.~Vishwanathan, and R.~Garnett (eds.), \emph{Advances in Neural Information Processing Systems}, volume~30. Curran Associates, Inc., 2017.

\bibitem[Ma \& Pellegrini(2019)Ma and Pellegrini]{10.1093/bioinformatics/btz592}
Feiyang Ma and Matteo Pellegrini.
\newblock {ACTINN: automated identification of cell types in single cell {RNA} sequencing}.
\newblock \emph{Bioinformatics}, 36\penalty0 (2):\penalty0 533--538, 07 2019.
\newblock ISSN 1367-4803.
\newblock \doi{10.1093/bioinformatics/btz592}.
\newblock URL \url{https://doi.org/10.1093/bioinformatics/btz592}.

\bibitem[Mai et~al.(2020)Mai, Shim, Jeong, Sanner, Kim, and Jang]{aser2020}
Zheda Mai, Dongsub Shim, Jihwan Jeong, Scott Sanner, Hyunwoo Kim, and Jongseong Jang.
\newblock Adversarial shapley value experience replay for task-free continual learning.
\newblock \emph{CoRR}, abs/2009.00093, 2020.
\newblock URL \url{https://arxiv.org/abs/2009.00093}.

\bibitem[Mathelin et~al.(2022)Mathelin, Deheeger, MOUGEOT, and Vayatis]{mathelin2022discrepancybased}
Antoine~De Mathelin, Fran{\c{c}}ois Deheeger, Mathilde MOUGEOT, and Nicolas Vayatis.
\newblock Discrepancy-based active learning for domain adaptation.
\newblock In \emph{International Conference on Learning Representations}, 2022.
\newblock URL \url{https://openreview.net/forum?id=p98WJxUC3Ca}.

\bibitem[Mayer \& Timofte(2020)Mayer and Timofte]{9093556}
C.~Mayer and R.~Timofte.
\newblock Adversarial sampling for active learning.
\newblock In \emph{2020 IEEE Winter Conference on Applications of Computer Vision (WACV)}, pp.\  3060--3068, Los Alamitos, CA, USA, mar 2020. IEEE Computer Society.
\newblock \doi{10.1109/WACV45572.2020.9093556}.
\newblock URL \url{https://doi.ieeecomputersociety.org/10.1109/WACV45572.2020.9093556}.

\bibitem[McClelland et~al.(1995)McClelland, McNaughton, and O'Reilly]{mcclelland1995there}
James~L McClelland, Bruce~L McNaughton, and Randall~C O'Reilly.
\newblock Why there are complementary learning systems in the hippocampus and neocortex: insights from the successes and failures of connectionist models of learning and memory.
\newblock \emph{Psychological review}, 102\penalty0 (3):\penalty0 419, 1995.

\bibitem[McCloskey \& Cohen(1989)McCloskey and Cohen]{mccloskey1989catastrophic}
Michael McCloskey and Neal~J Cohen.
\newblock Catastrophic interference in connectionist networks: The sequential learning problem.
\newblock In \emph{Psychology of learning and motivation}, volume~24, pp.\  109--165. Elsevier, 1989.

\bibitem[Mermillod et~al.(2013)Mermillod, Bugaiska, and BONIN]{stabilityplasticity}
Martial Mermillod, Aurélia Bugaiska, and Patrick BONIN.
\newblock The stability-plasticity dilemma: investigating the continuum from catastrophic forgetting to age-limited learning effects.
\newblock \emph{Frontiers in Psychology}, 4, 2013.
\newblock ISSN 1664-1078.
\newblock \doi{10.3389/fpsyg.2013.00504}.
\newblock URL \url{https://www.frontiersin.org/article/10.3389/fpsyg.2013.00504}.

\bibitem[Minoux(1978)]{minoux1978accelerated}
M.~Minoux.
\newblock Accelerated greedy algorithms for maximizing submodular set functions.
\newblock In \emph{Optimization Techniques}, 1978.

\bibitem[Mirzasoleiman et~al.(2015)Mirzasoleiman, Badanidiyuru, Karbasi, Vondr{\'a}k, and Krause]{mirzasoleiman2015lazier}
Baharan Mirzasoleiman, Ashwinkumar Badanidiyuru, Amin Karbasi, Jan Vondr{\'a}k, and Andreas Krause.
\newblock Lazier than lazy greedy.
\newblock In \emph{Proceedings of the AAAI Conference on Artificial Intelligence}, volume~29, 2015.

\bibitem[Perkonigg et~al.(2021)Perkonigg, Hofmanninger, and Langs]{Perkonigg2021-ee}
Matthias Perkonigg, Johannes Hofmanninger, and Georg Langs.
\newblock Continual active learning for efficient adaptation of machine learning models to changing image acquisition.
\newblock In \emph{Lecture Notes in Computer Science}, Lecture notes in computer science, pp.\  649--660. Springer International Publishing, Cham, 2021.

\bibitem[Pinsler et~al.(2019)Pinsler, Gordon, Nalisnick, and Hern\'{a}ndez-Lobato]{NEURIPS2019_84c2d486}
Robert Pinsler, Jonathan Gordon, Eric Nalisnick, and Jos\'{e}~Miguel Hern\'{a}ndez-Lobato.
\newblock Bayesian batch active learning as sparse subset approximation.
\newblock In H.~Wallach, H.~Larochelle, A.~Beygelzimer, F.~d\textquotesingle Alch\'{e}-Buc, E.~Fox, and R.~Garnett (eds.), \emph{Advances in Neural Information Processing Systems}, volume~32. Curran Associates, Inc., 2019.
\newblock URL \url{https://proceedings.neurips.cc/paper/2019/file/84c2d4860a0fc27bcf854c444fb8b400-Paper.pdf}.

\bibitem[Ratcliff(1990)]{Ratcliff1990}
R~Ratcliff.
\newblock Connectionist models of recognition memory: constraints imposed by learning and forgetting functions.
\newblock \emph{Psychol Rev}, 97\penalty0 (2):\penalty0 285--308, April 1990.

\bibitem[Riemer et~al.(2018)Riemer, Cases, Ajemian, Liu, Rish, Tu, and Tesauro]{MER2018}
Matthew Riemer, Ignacio Cases, Robert Ajemian, Miao Liu, Irina Rish, Yuhai Tu, and Gerald Tesauro.
\newblock Learning to learn without forgetting by maximizing transfer and minimizing interference.
\newblock \emph{CoRR}, abs/1810.11910, 2018.

\bibitem[Robins(1995)]{Robins95}
Anthony Robins.
\newblock Catastrophic forgetting, rehearsal and pseudorehearsal.
\newblock \emph{Connection Science}, 7\penalty0 (2):\penalty0 123--146, 1995.
\newblock \doi{10.1080/09540099550039318}.
\newblock URL \url{https://doi.org/10.1080/09540099550039318}.

\bibitem[Rozen et~al.(2019)Rozen, Shwartz, Aharoni, and Dagan]{rozen-etal-2019-diversify}
Ohad Rozen, Vered Shwartz, Roee Aharoni, and Ido Dagan.
\newblock Diversify your datasets: Analyzing generalization via controlled variance in adversarial datasets.
\newblock In \emph{Proceedings of the 23rd Conference on Computational Natural Language Learning (CoNLL)}, pp.\  196--205, Hong Kong, China, November 2019. Association for Computational Linguistics.
\newblock \doi{10.18653/v1/K19-1019}.
\newblock URL \url{https://aclanthology.org/K19-1019}.

\bibitem[Schwartz et~al.(2020)Schwartz, Dodge, Smith, and Etzioni]{greenai2020}
Roy Schwartz, Jesse Dodge, Noah~A. Smith, and Oren Etzioni.
\newblock Green ai.
\newblock \emph{Commun. ACM}, 63\penalty0 (12):\penalty0 54–63, nov 2020.
\newblock ISSN 0001-0782.
\newblock \doi{10.1145/3381831}.
\newblock URL \url{https://doi.org/10.1145/3381831}.

\bibitem[Schwenk et~al.(2017)Schwenk, Barrault, Conneau, and LeCun]{Schwenk2017VeryDC}
Holger Schwenk, Lo{\"i}c Barrault, Alexis Conneau, and Yann LeCun.
\newblock Very deep convolutional networks for text classification.
\newblock In \emph{EACL}, 2017.

\bibitem[Sener \& Savarese(2017)Sener and Savarese]{https://doi.org/10.48550/arxiv.1708.00489}
Ozan Sener and Silvio Savarese.
\newblock Active learning for convolutional neural networks: A core-set approach, 2017.
\newblock URL \url{https://arxiv.org/abs/1708.00489}.

\bibitem[Sener \& Savarese(2018)Sener and Savarese]{sener2018active}
Ozan Sener and Silvio Savarese.
\newblock Active learning for convolutional neural networks: A core-set approach.
\newblock In \emph{International Conference on Learning Representations}, 2018.
\newblock URL \url{https://openreview.net/forum?id=H1aIuk-RW}.

\bibitem[Senzaki \& Hamelain(2021)Senzaki and Hamelain]{edgeAL}
Yuya Senzaki and Christian Hamelain.
\newblock Active learning for deep neural networks on edge devices, 2021.
\newblock URL \url{https://arxiv.org/abs/2106.10836}.

\bibitem[Settles()]{settles2012}
Burr Settles.
\newblock Active learning.
\newblock Morgan \& Claypool Publishers, 2012.

\bibitem[Settles(2009)]{settles2009active}
Burr Settles.
\newblock Active learning literature survey.
\newblock Computer Sciences Technical Report 1648, University of Wisconsin--Madison, 2009.
\newblock URL \url{http://axon.cs.byu.edu/~martinez/classes/778/Papers/settles.activelearning.pdf}.

\bibitem[Settles(2011)]{settles2010}
Burr Settles.
\newblock From theories to queries: Active learning in practice.
\newblock In Isabelle Guyon, Gavin Cawley, Gideon Dror, Vincent Lemaire, and Alexander Statnikov (eds.), \emph{Active Learning and Experimental Design workshop In conjunction with AISTATS 2010}, volume~16 of \emph{Proceedings of Machine Learning Research}, pp.\  1--18, Sardinia, Italy, 16 May 2011. PMLR.
\newblock URL \url{https://proceedings.mlr.press/v16/settles11a.html}.

\bibitem[Shoeybi et~al.(2019)Shoeybi, Patwary, Puri, LeGresley, Casper, and Catanzaro]{megatron}
Mohammad Shoeybi, Mostofa Patwary, Raul Puri, Patrick LeGresley, Jared Casper, and Bryan Catanzaro.
\newblock Megatron-lm: Training multi-billion parameter language models using model parallelism.
\newblock \emph{CoRR}, abs/1909.08053, 2019.
\newblock URL \url{http://arxiv.org/abs/1909.08053}.

\bibitem[Shui et~al.(2019)Shui, Zhou, Gagn{\'{e}}, and Wang]{discreAISTATS}
Changjian Shui, Fan Zhou, Christian Gagn{\'{e}}, and Boyu Wang.
\newblock Deep active learning: Unified and principled method for query and training.
\newblock \emph{CoRR}, abs/1911.09162, 2019.
\newblock URL \url{http://arxiv.org/abs/1911.09162}.

\bibitem[Titsias et~al.(2020)Titsias, Schwarz, de~G.~Matthews, Pascanu, and Teh]{Titsias2020Functional}
Michalis~K. Titsias, Jonathan Schwarz, Alexander~G. de~G.~Matthews, Razvan Pascanu, and Yee~Whye Teh.
\newblock Functional regularisation for continual learning with gaussian processes.
\newblock In \emph{International Conference on Learning Representations}, 2020.
\newblock URL \url{https://openreview.net/forum?id=HkxCzeHFDB}.

\bibitem[Tschandl et~al.(2018)Tschandl, Rosendahl, and Kittler]{Tschandl2018TheHD}
Philipp Tschandl, Cliff Rosendahl, and Harald Kittler.
\newblock The ham10000 dataset, a large collection of multi-source dermatoscopic images of common pigmented skin lesions.
\newblock \emph{Scientific Data}, 5, 2018.

\bibitem[Wallingford et~al.(2023)Wallingford, Kusupati, Alizadeh-Vahid, Walsman, Kembhavi, and Farhadi]{wallingford2023fluid}
Matthew Wallingford, Aditya Kusupati, Keivan Alizadeh-Vahid, Aaron Walsman, Aniruddha Kembhavi, and Ali Farhadi.
\newblock {FLUID}: A unified evaluation framework for flexible sequential data.
\newblock \emph{Transactions on Machine Learning Research}, 2023.
\newblock ISSN 2835-8856.
\newblock URL \url{https://openreview.net/forum?id=UvJBKWaSSH}.

\bibitem[Warstadt et~al.(2018)Warstadt, Singh, and Bowman]{warstadt2018neural}
Alex Warstadt, Amanpreet Singh, and Samuel~R Bowman.
\newblock Neural network acceptability judgments.
\newblock \emph{arXiv preprint arXiv:1805.12471}, 2018.

\bibitem[Wei et~al.(2015)Wei, Iyer, and Bilmes]{fass2015}
Kai Wei, Rishabh Iyer, and Jeff Bilmes.
\newblock Submodularity in data subset selection and active learning.
\newblock In \emph{Proceedings of the 32nd International Conference on International Conference on Machine Learning - Volume 37}, ICML'15, pp.\  1954–1963. JMLR.org, 2015.

\bibitem[Wolf et~al.(2020)Wolf, Debut, Sanh, Chaumond, Delangue, Moi, Cistac, Rault, Louf, Funtowicz, Davison, Shleifer, von Platen, Ma, Jernite, Plu, Xu, Scao, Gugger, Drame, Lhoest, and Rush]{wolf-etal-2020-transformers}
Thomas Wolf, Lysandre Debut, Victor Sanh, Julien Chaumond, Clement Delangue, Anthony Moi, Pierric Cistac, Tim Rault, R\'{e}mi Louf, Morgan Funtowicz, Joe Davison, Sam Shleifer, Patrick von Platen, Clara Ma, Yacine Jernite, Julien Plu, Canwen Xu, Teven~Le Scao, Sylvain Gugger, Mariama Drame, Quentin Lhoest, and Alexander~M. Rush.
\newblock Transformers: State-of-the-art natural language processing.
\newblock In \emph{Proceedings of the 2020 Conference on Empirical Methods in Natural Language Processing: System Demonstrations}, pp.\  38--45, Online, October 2020. Association for Computational Linguistics.
\newblock URL \url{https://www.aclweb.org/anthology/2020.emnlp-demos.6}.

\bibitem[Xiao et~al.(2017)Xiao, Rasul, and Vollgraf]{fmnist}
Han Xiao, Kashif Rasul, and Roland Vollgraf.
\newblock Fashion-mnist: a novel image dataset for benchmarking machine learning algorithms.
\newblock \emph{CoRR}, abs/1708.07747, 2017.
\newblock URL \url{http://arxiv.org/abs/1708.07747}.

\bibitem[Xie et~al.(2021)Xie, Jiang, Mora, and Li]{XIE20215874}
Bingbing Xie, Qin Jiang, Antonio Mora, and Xuri Li.
\newblock Automatic cell type identification methods for single-cell {RNA} sequencing.
\newblock \emph{Computational and Structural Biotechnology Journal}, 19:\penalty0 5874--5887, 2021.
\newblock ISSN 2001-0370.
\newblock \doi{https://doi.org/10.1016/j.csbj.2021.10.027}.
\newblock URL \url{https://www.sciencedirect.com/science/article/pii/S2001037021004499}.

\bibitem[Yang et~al.(2021{\natexlab{a}})Yang, Shi, and Ni]{medmnistv1}
Jiancheng Yang, Rui Shi, and Bingbing Ni.
\newblock Medmnist classification decathlon: A lightweight automl benchmark for medical image analysis.
\newblock In \emph{IEEE 18th International Symposium on Biomedical Imaging (ISBI)}, pp.\  191--195, 2021{\natexlab{a}}.

\bibitem[Yang et~al.(2021{\natexlab{b}})Yang, Shi, Wei, Liu, Zhao, Ke, Pfister, and Ni]{medmnistv2}
Jiancheng Yang, Rui Shi, Donglai Wei, Zequan Liu, Lin Zhao, Bilian Ke, Hanspeter Pfister, and Bingbing Ni.
\newblock Medmnist v2: A large-scale lightweight benchmark for 2d and 3d biomedical image classification.
\newblock \emph{arXiv preprint arXiv:2110.14795}, 2021{\natexlab{b}}.

\bibitem[Yoo \& Kweon(2019)Yoo and Kweon]{DBLP:journals/corr/abs-1905-03677}
Donggeun Yoo and In~So Kweon.
\newblock Learning loss for active learning.
\newblock \emph{CoRR}, abs/1905.03677, 2019.
\newblock URL \url{http://arxiv.org/abs/1905.03677}.

\bibitem[Zhang et~al.(2017)Zhang, Stafman, Or, and Freedman]{slaq2017}
Haoyu Zhang, Logan Stafman, Andrew Or, and Michael~J. Freedman.
\newblock Slaq: Quality-driven scheduling for distributed machine learning.
\newblock SoCC '17, pp.\  390–404, New York, NY, USA, 2017. Association for Computing Machinery.
\newblock ISBN 9781450350280.
\newblock \doi{10.1145/3127479.3127490}.
\newblock URL \url{https://doi.org/10.1145/3127479.3127490}.

\bibitem[Zhang et~al.(2023)Zhang, Chen, Canal, Mussmann, Zhu, Du, Jamieson, and Nowak]{zhang2023labelbench}
Jifan Zhang, Yifang Chen, Gregory Canal, Stephen Mussmann, Yinglun Zhu, Simon~Shaolei Du, Kevin Jamieson, and Robert~D Nowak.
\newblock Labelbench: A comprehensive framework for benchmarking label-efficient learning, 2023.

\bibitem[Zhang et~al.(2015)Zhang, Zhao, and LeCun]{DBLP:journals/corr/ZhangZL15}
Xiang Zhang, Junbo~Jake Zhao, and Yann LeCun.
\newblock Character-level convolutional networks for text classification.
\newblock \emph{CoRR}, abs/1509.01626, 2015.
\newblock URL \url{http://arxiv.org/abs/1509.01626}.

\bibitem[Zheng et~al.(2022)Zheng, Li, Zhang, Zhuang, Chen, Huang, Wang, Xu, Zhuo, Gonzalez, and Stoica]{alpa2022}
Lianmin Zheng, Zhuohan Li, Hao Zhang, Yonghao Zhuang, Zhifeng Chen, Yanping Huang, Yida Wang, Yuanzhong Xu, Danyang Zhuo, Joseph~E. Gonzalez, and Ion Stoica.
\newblock Alpa: Automating inter- and intra-operator parallelism for distributed deep learning.
\newblock \emph{CoRR}, abs/2201.12023, 2022.
\newblock URL \url{https://arxiv.org/abs/2201.12023}.

\bibitem[Zhou et~al.(2021)Zhou, Shin, Gurudu, Gotway, and Liang]{ZHOU2021101997}
Zongwei Zhou, Jae~Y. Shin, Suryakanth~R. Gurudu, Michael~B. Gotway, and Jianming Liang.
\newblock Active, continual fine tuning of convolutional neural networks for reducing annotation efforts.
\newblock \emph{Medical Image Analysis}, 71:\penalty0 101997, 2021.
\newblock ISSN 1361-8415.
\newblock \doi{https://doi.org/10.1016/j.media.2021.101997}.
\newblock URL \url{https://www.sciencedirect.com/science/article/pii/S1361841521000438}.

\bibitem[Zhu \& Bento(2017)Zhu and Bento]{DBLP:journals/corr/ZhuB17}
Jia{-}Jie Zhu and Jos{\'{e}} Bento.
\newblock Generative adversarial active learning.
\newblock \emph{CoRR}, abs/1702.07956, 2017.
\newblock URL \url{http://arxiv.org/abs/1702.07956}.

\end{thebibliography}
\bibliographystyle{tmlr}

\appendix
\onecolumn
\section{Additional Experimental Details on Main Results}
\subsection{Procedure for normalized accuracy plots}
Here we again explain the procedure to generate the normalized accuracy versus speedup plots, as reported in figures~\ref{fig:main_result}, \ref{fig:all_budget}, and \ref{fig:all_datasets}. These plots help us to understand and compare the performance of different CAL methods against the baseline AL. For every dataset, we first get the respective accuracies and speedups of CAL methods and baseline AL, for every budget. This is also reported as a tabular form besides the scatter plot visualizations in section~\ref{sec: Results in Tabular Form}. 

For \textbf{every budget}, we normalize the accuracies of each method by that of baseline AL. This makes the baseline always at \textbf{1}, irrespective of budget and dataset. Relative performances greater than 1 indicate better than baseline accuracy (and similarly for the less than 1 case). Note that, for CIFAR10 and Amazon polarity review, budgets go from 10\% to 50\% in increments of 10\%. However, for FMNIST, MedMNIST, and Celltype datasets, it goes from 10\% to 30\% in increments of 5\%. Lastly for COLA, the budgets are different from the above, therefore, we don't include that when we average across the dataset at a fixed budget. Having said that, 

\begin{itemize}
    \item Keeping the budgets fixed to 10\%, 20\%, and 30\% and averaging over the datasets (except COLA, since it has a different budget) will give us figure~\ref{fig:all_budget}.
    \item Keeping the dataset fixed, averaging the relative accuracy v.s.\  speedups across different budgets will give us figure~\ref{fig:all_datasets}.  
    \item Further averaging the above across different datasets will give us the main result figure~\ref{fig:main_result}.  
\end{itemize}

From the results above, we can infer that CAL-DER and specialized methods such as CAL-SD/SDS2 always provide a speedup, all while preserving the accuracy, if not better.  

\subsection{Results in Tabular Form}    
\label{sec: Results in Tabular Form}
In this section, we expand our results mentioned in section~\ref{sec:performance_vs_speedup}. In particular, we report the absolute accuracies for each dataset, at every budget, and plot it against the observed speedup. All methods highlighted in \textcolor{blue}{blue} are methods that use CAL. Note that all the results in this section are for uncertainty-based query pool acquisition functions. The choice of budget scale is taken from the previous works \citep{distil2021, ein-dor-etal-2020-active}. Each dataset has a different complexity (and we train on them using different neural architectures), therefore they differ in the labeling budget. 

\begin{figure}[H]
    \centering
\includegraphics[width=\linewidth]{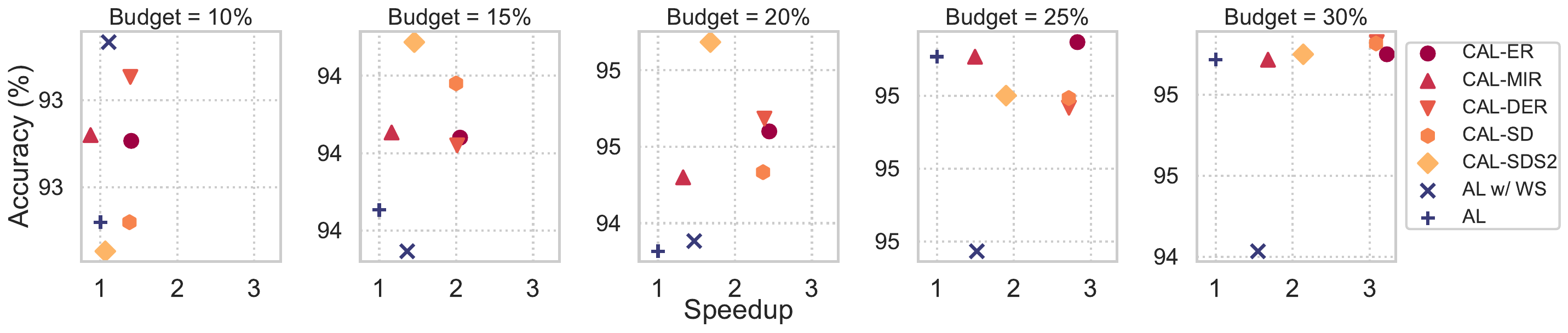} 
    \caption{FMNIST Results}
    \label{fig:fmnist}
\end{figure}

\paragraph{FMNIST} Please refer to table~\ref{tab:FMNIST_uncert} and figure~\ref{fig:fmnist}.

\begin{small}
\begin{table}[H]
\centering
\scalebox{0.85}{
\begin{tabular}{c|ccccc|ccccc} 
 \toprule[1.5pt]
                 \multicolumn{1}{c|}{} &\multicolumn{5}{c|}{\textbf{Test Accuracy (\%)}}                              & \multicolumn{5}{c}{\textbf{Factor Speedup}}                                   \\ \midrule
 \textbf{Method}  & \textbf{10\%} & \textbf{15\%} & \textbf{20\%} & \textbf{25\%} & \textbf{30\%} & \textbf{10\%} & \textbf{15\%} & \textbf{20\%} & \textbf{25\%} & \textbf{30\%} \\ \midrule 
     \textcolor{blue}{CAL-ER} & 92.6 {\tiny $\pm$ 0.1} & 93.9 {\tiny $\pm$ 0.2} & 94.5 {\tiny $\pm$ 0.1} & \textbf{94.9} {\tiny $\pm$ 0.2} & \textbf{94.9} {\tiny $\pm$ 0.2} & 1.5$\times$ & 1.4$\times$ & 2.0$\times$  & 2.4$\times$ & 2.8 $\times$     \\
\textcolor{blue}{CAL-MIR} & 92.6 {\tiny $\pm$ 0.3} & 93.9 {\tiny $\pm$ 0.2} & 94.5 {\tiny $\pm$ 0.0} & \textbf{94.9} {\tiny $\pm$ 0.1} & \textbf{94.9} {\tiny $\pm$ 0.0} & 0.9 $\times$ & 1.2$\times$ & 1.3$\times$ & 1.5$\times$ & 1.7$\times$ \\
\textcolor{blue}{CAL-DER} & \textbf{92.7} {\tiny $\pm$ 0.1} & 93.9 {\tiny $\pm$ 0.1} & 94.5 {\tiny $\pm$ 0.1} & 94.8 {\tiny $\pm$ 0.2} & \textbf{94.9} {\tiny $\pm$ 0.1} & 1.4 $\times$ & 2.0$\times$ & 2.4$\times$ & 2.7$\times$ & 3.1$\times$ \\
\textcolor{blue}{CAL-SD} & 92.6 {\tiny $\pm$ 0.1} & \textbf{94.0} {\tiny $\pm$ 0.2} & 94.5 {\tiny $\pm$ 0.1} & 94.8 {\tiny $\pm$ 0.2} & \textbf{94.9} {\tiny $\pm$ 0.1} & 1.4 $\times$ & 2.0$\times$ & 2.4$\times$ & 2.7$\times$ & 3.1$\times$            \\
\textcolor{blue}{CAL-SDS2} & 92.6 {\tiny $\pm$ 0.1} & \textbf{94.0} {\tiny $\pm$ 0.2} & \textbf{94.6} {\tiny $\pm$ 0.2} & \textbf{94.9} {\tiny $\pm$ 0.1} & \textbf{94.9} {\tiny $\pm$ 0.1} & 1.1$\times$ & 1.5$\times$ & 1.7$\times$ & 1.9$\times$ & 2.1$\times$ \\ \midrule
AL w/ WS & \textbf{92.7} {\tiny $\pm$ 0.3} & 93.8 {\tiny $\pm$ 0.2} & 94.4 {\tiny $\pm$ 0.1} & 94.6 {\tiny $\pm$ 0.1} & 94.4 {\tiny $\pm$ 0.2}  & 1.1$\times$ & 1.4$\times$ & 1.5$\times$ & 1.5$\times$ & 1.5$\times$ \\
AL & 92.6 {\tiny $\pm$ 0.3} & 93.8 {\tiny $\pm$ 0.0} & 94.4 {\tiny $\pm$ 0.1} & \textbf{94.9} {\tiny $\pm$ 0.2} & \textbf{94.9} {\tiny $\pm$ 0.1} & 1.0$ \times$ & 1.0$\times$ & 1.0$\times$ & 1.0$\times$ & 1.0$\times$ \\ \midrule

\end{tabular}}
  \caption{FMNIST Results}
  \label{tab:FMNIST_uncert}
\end{table}
\end{small}

\begin{figure}[H]
    \centering
    \includegraphics[width=\linewidth]{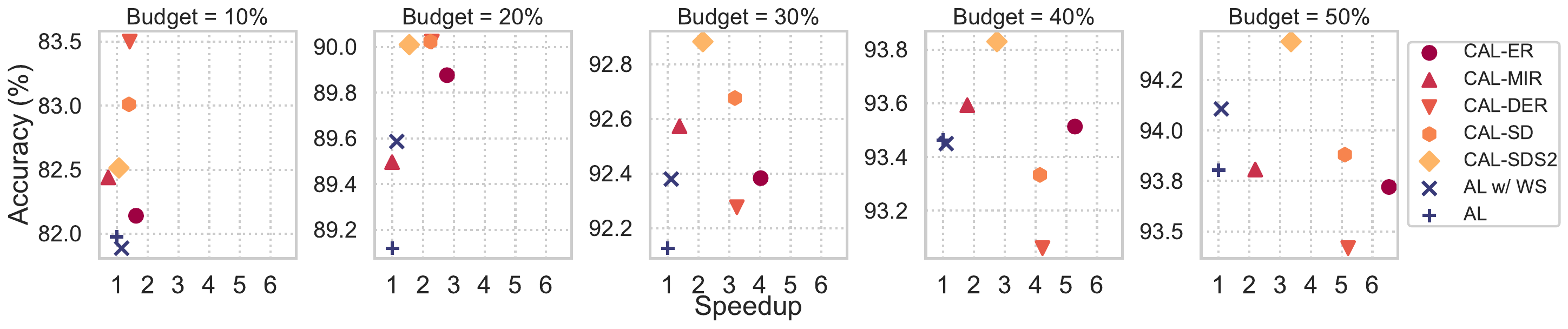} 
    \caption{CIFAR-10 Results}
    \label{fig:cifar}
\end{figure}

\paragraph{CIFAR10} Please refer to table~\ref{tab:CIFAR_uncert} and figure~\ref{fig:cifar}.

\begin{table}[H]
\centering
\scalebox{0.85}{
\begin{tabular}{c|ccccc|ccccc} 
 \toprule[1.5pt]
\multicolumn{1}{c|}{} &\multicolumn{5}{c|}{\textbf{Test Accuracy (\%)}}                              & \multicolumn{5}{c}{\textbf{Factor Speedup}}                                   \\ \midrule
\textbf{Method}  & \textbf{10\%} & \textbf{20\%} & \textbf{30\%} & \textbf{40\%} & \textbf{50\%} & \textbf{10\%} & \textbf{20\%} & \textbf{30\%} & \textbf{40\%} & \textbf{50\%} \\ \midrule 
\textcolor{blue}{CAL-ER} & 82.1 {\tiny $\pm$ 0.5} & 89.9 {\tiny $\pm$ 0.3} & 92.4 {\tiny $\pm$ 0.1} & 93.5 {\tiny $\pm$ 0.1} & 93.7 {\tiny $\pm$ 0.3} & 1.6$\times$ & 2.8$\times$ & 4.0$\times$  & 5.3$\times$ & 6.5$\times$           \\
\textcolor{blue}{CAL-MIR} & 82.4 {\tiny $\pm$ 0.4} & 89.5 {\tiny $\pm$ 0.3}	& 92.6 {\tiny $\pm$ 0.3} & 93.6 {\tiny $\pm$ 0.1} & 93.8 {\tiny $\pm$ 0.2} & 0.7 $\times$ & 1.0$\times$ & 1.4$\times$ & 1.8$\times$ & 2.2$\times$ \\
\textcolor{blue}{CAL-DER} & \textbf{83.5} {\tiny $\pm$ 0.1} & 90.0 {\tiny $\pm$ 0.4} & 92.3 {\tiny $\pm$ 0.1} & 93.1 {\tiny $\pm$ 0.2} & 93.4 {\tiny $\pm$ 0.1} & 1.4$\times$ & 2.3$\times$ & 3.2$\times$ & 4.2$\times$ & 5.2$\times$ \\
\textcolor{blue}{CAL-SD} & 83.0 {\tiny $\pm$ 0.0} & 90.0 {\tiny $\pm$ 0.4} & 92.7 {\tiny $\pm$ 0.2} & 93.3 {\tiny $\pm$ 0.3} & 93.9 {\tiny $\pm$ 0.3} & 1.4$\times$ & 2.2$\times$ & 3.2$\times$ & 4.1$\times$ & 5.1$\times$           \\
\textcolor{blue}{CAL-SDS2} & 82.5 {\tiny $\pm$ 0.1} & \textbf{90.1} {\tiny $\pm$ 0.2} & \textbf{92.9}  {\tiny $\pm$ 0.4} & \textbf{94.0} {\tiny $\pm$ 0.2} & \textbf{94.4} {\tiny $\pm$ 0.1} & 1.1$\times$ & 1.6$\times$ & 2.1$\times$ & 2.7$\times$ & 3.4$\times$ \\
\midrule
AL w/ WS & 81.9 {\tiny $\pm$ 0.4} & 89.6 {\tiny $\pm$ 0.5} & 92.4 {\tiny $\pm$ 0.2} & 93.5 {\tiny $\pm$ 0.1} & 94.1 {\tiny $\pm$ 0.1} & 1.2$\times$ & 1.1$\times$ & 1.1$\times$ & 1.1$\times$ & 1.1$\times$ \\
AL & 82.0 {\tiny $\pm$ 0.3} & 89.1 {\tiny $\pm$ 0.2} & 92.1 {\tiny $\pm$ 0.4} & 93.5 {\tiny $\pm$ 0.3} & 93.8 {\tiny $\pm$ 0.2} & 1.0$\times$ & 1.0$\times$ & 1.0$\times$ & 1.0$\times$ & 1.0$\times$ \\ \midrule

\end{tabular}}
  \caption{CIFAR-10 Results}
  \label{tab:CIFAR_uncert}
\end{table}

\begin{figure}[H]
    \centering
    \includegraphics[width=\linewidth]{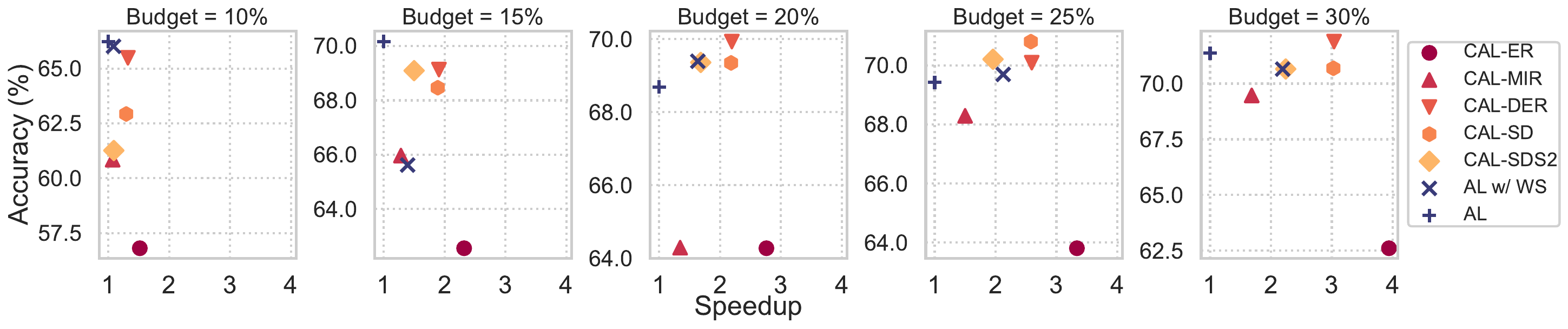} 
    \caption{MedMNIST Results}
    \label{fig:medmnist}
\end{figure}

\paragraph{MedMNIST} Please refer to table~\ref{tab:MedMNIST_uncert} and figure~\ref{fig:medmnist}.

\begin{table}[H]
\centering
\scalebox{0.85}{
\begin{tabular}{c|ccccc|ccccc} 
 \toprule[1.5pt]
                 \multicolumn{1}{c|}{} & \multicolumn{5}{c|}{\textbf{Test Accuracy (\%)}}                              & \multicolumn{5}{c}{\textbf{Factor Speedup}}                                   \\ \midrule
 \textbf{Method}  & \textbf{10\%} & \textbf{15\%} & \textbf{20\%} & \textbf{25\%} & \textbf{30\%} & \textbf{10\%} & \textbf{15\%} & \textbf{20\%} & \textbf{25\%} & \textbf{30\%} \\ \midrule

 \textcolor{blue}{CAL-ER} & 64.7 {\tiny $\pm$ 4.4} & 63.1 {\tiny $\pm$ 5.2} & 68.4 {\tiny $\pm$ 1.2} & 66.1 {\tiny $\pm$ 3.7} & 67.5 {\tiny $\pm$ 2.2} & 1.4$\times$ &	2.0$\times$ &	2.4$\times$ & 2.9$\times$ &	3.4$\times$    \\
 
\textcolor{blue}{CAL-MIR} & 64.1 {\tiny $\pm$ 2.8} & 66.2 {\tiny $\pm$ 1.0} & 69.2 {\tiny $\pm$ 1.7} & 70.7 {\tiny $\pm$ 1.1} & 71.8 {\tiny $\pm$ 0.8} & 1.0$\times$ &	1.2$\times$ &	1.4$\times$ &	1.6$\times$ & 1.9$\times$ \\

\textcolor{blue}{CAL-DER} & 67.1 {\tiny $\pm$ 2.9} & 67.9 {\tiny $\pm$ 2.0} & \textbf{69.8} {\tiny $\pm$ 1.2} & 71.5 {\tiny $\pm$ 0.6} & 72.2 {\tiny $\pm$ 0.6} & 1.2$\times$ &	1.6$\times$ &	2.1$\times$ &	2.6$\times$ &	3.2$\times$\\

\textcolor{blue}{CAL-SD} & 65.1 {\tiny $\pm$ 3.1} & 67.8 {\tiny $\pm$ 2.3} & 70.6 {\tiny $\pm$ 0.9} & 71.1 {\tiny $\pm$ 1.1} & 72.1 {\tiny $\pm$ 1.1} & 1.2$\times$ &	1.7$\times$ &	2.0$\times$ &	2.5$\times$ &	2.9$\times$      \\

\textcolor{blue}{CAL-SDS2} & 66.1 {\tiny $\pm$ 3.6}&	68.1 {\tiny $\pm$ 3.0}&	\textbf{69.8} {\tiny $\pm$ 1.1}&	\textbf{71.6} {\tiny $\pm$ 1.6} &	\textbf{72.5} {\tiny $\pm$ 1.6} & 1.1$\times$ &	1.5$\times$ &	1.7$\times$ &	2.0$\times$ &	2.2$\times$ \\
\midrule

AL w/ WS & 67.0	{\tiny $\pm$ 1.4}& 67.5 {\tiny $\pm$ 0.7} &	69.5 {\tiny $\pm$ 0.6} & 70.3 {\tiny $\pm$ 1.0}&	71.3 {\tiny $\pm$ 1.0}& 1.1$\times$ &	1.3$\times$ &	1.5$\times$ &	1.7$\times$ &	2.0$\times$ \\

AL & \textbf{67.2} {\tiny $\pm$ 0.7} & \textbf{68.3}  {\tiny $\pm $1.3}& 68.8 {\tiny $\pm$ 0.8}&	69.0 {\tiny $\pm$ 2.0} & 70.9 {\tiny $\pm$ 1.5} & 1.0$\times$ & 1.0$\times$ & 1.0$\times$ & 1.0$\times$ & 1.0$\times$ \\ \midrule

\end{tabular}
  }
  \caption{MedMNIST Results}
  \label{tab:MedMNIST_uncert}
\end{table}

\begin{figure}[H]
    \centering
    \includegraphics[width=1.0\linewidth]{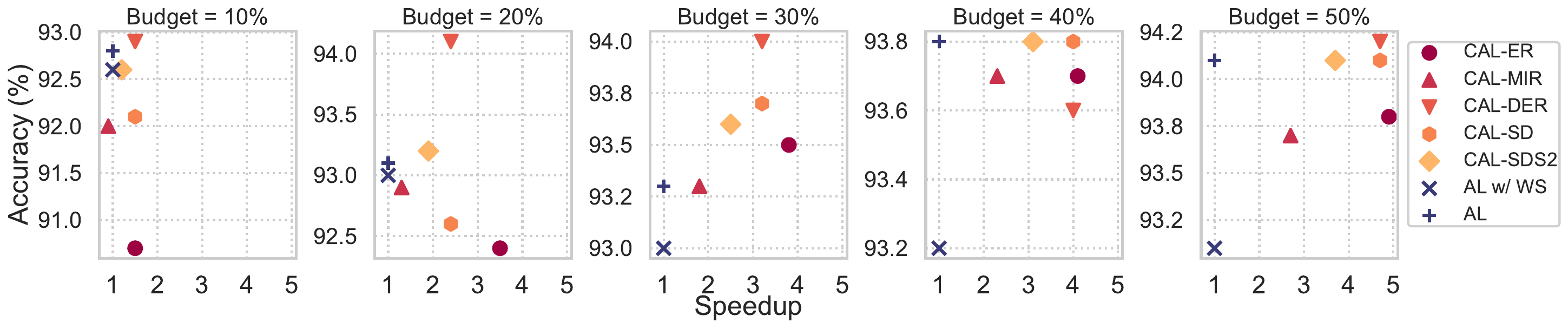} 
    \caption{Amazon Polarity Results}
    \label{fig:amazonpo}
\end{figure}

\paragraph{Amazon Polarity Review} Please refer to table~\ref{tab:Amazon_uncert} and figure~\ref{fig:amazonpo}.

\begin{table}[H]
\centering
\scalebox{0.85}{
\begin{tabular}{c|ccccc|ccccc} 
 \toprule[1.5pt]
                 \multicolumn{1}{c|}{} & \multicolumn{5}{c|}{\textbf{Test Accuracy (\%)}}                              & \multicolumn{5}{c}{\textbf{Factor Speedup}}                                   \\ \midrule
 \textbf{Method}  & \textbf{10\%} & \textbf{20\%} & \textbf{30\%} & \textbf{40\%} & \textbf{50\%} & \textbf{10\%} & \textbf{20\%} & \textbf{30\%} & \textbf{40\%} & \textbf{50\%} \\ \midrule 
\textcolor{blue}{CAL-ER}  & 90.7 {\tiny $\pm$ 3.1}               & 92.4 {\tiny $\pm$ 1.2} & 93.5 {\tiny $\pm$ 0.1}& 93.7 {\tiny $\pm$ 0.2} & 93.8 {\tiny $\pm$ 0.2} & 1.5x            & 3.5x & 3.3x & 4.1x & 4.9x \\
\textcolor{blue}{CAL-MIR} & 92.0  {\tiny $\pm$ 0.9}              & 92.9 {\tiny $\pm$ 0.1} & 93.3 {\tiny $\pm$ 0.3}& 93.7 {\tiny $\pm$ 0.1}& 93.7 {\tiny $\pm$ 0.2} & 0.9x           & 1.3x & 1.8x & 2.3x & 2.7x \\
\textcolor{blue}{CAL-DER} & \textbf{92.9} {\tiny $\pm$ 0.3}               & \textbf{94.1} {\tiny $\pm$ 0.3} & \textbf{94.0} {\tiny $\pm$ 0.7}& 93.6 {\tiny $\pm$ 0.8} & \textbf{94.2} {\tiny $\pm$ 0.3} & 1.5x            & 2.4x & 3.2x & 4.0x & 4.7x \\
\textcolor{blue}{CAL-SD} & 92.1  {\tiny $\pm$ 0.3}              & 92.6 {\tiny $\pm$ 0.4}& 93.7 {\tiny $\pm$ 0.1}& \textbf{93.8} {\tiny $\pm$ 0.1}& 94.1 {\tiny $\pm$ 0.1} & 1.5x         & 2.4x & 3.2x & 4.0x & 4.7x \\
\textcolor{blue}{CAL-SDS2} & 92.6 {\tiny $\pm$ 0.3}               & 93.2 {\tiny $\pm$ 0.1} & 93.6 {\tiny $\pm$ 0.1} & \textbf{93.8} {\tiny $\pm$ 0.4}& 94.1 {\tiny $\pm$ 0.0} & 1.2x            & 1.9x & 2.5x & 3.1x & 3.7x \\
\midrule 
AL w/ WS & 92.6 {\tiny $\pm$ 0.5}                & 93.0 {\tiny $\pm$ 0.2}  & 93.0 {\tiny $\pm$ 0.1} & 93.2 {\tiny $\pm$ 0.3}  & 93.1 {\tiny $\pm$ 0.1}  & 1.0x            & 1.0x & 1.0x & 1.0x & 1.0x \\
AL & 92.8  {\tiny $\pm$ 0.2}               & 93.1 {\tiny $\pm$ 0.7}  & 93.3 {\tiny $\pm$ 1.1}  & \textbf{93.8} {\tiny $\pm$ 0.5}  & 94.1 {\tiny $\pm$ 0.2}  & 1.0x           & 1.0x & 1.0x & 1.0x & 1.0x\\ \midrule

\end{tabular}
  }
  \caption{Amazon Polarity Results}
  \label{tab:Amazon_uncert}
\end{table}

\begin{figure}[H]
    \centering
    \includegraphics[width=1.0\linewidth]{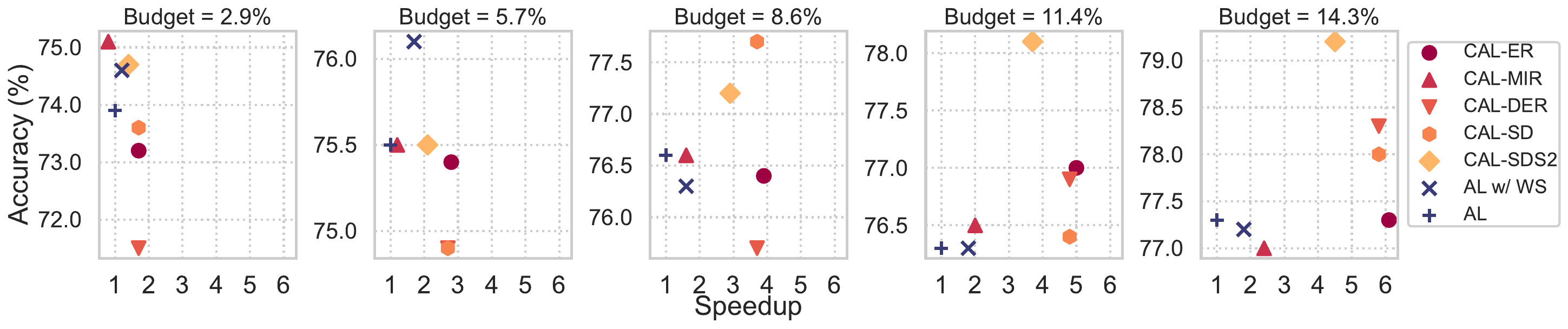} 
    \caption{COLA Results}
    \label{fig:cola}
\end{figure}

\paragraph{COLA} Please refer to table~\ref{tab:cola_uncert} and figure~\ref{fig:cola}.

\begin{table}[H]
\centering
\scalebox{0.85}{
\begin{tabular}{c|ccccc|ccccc} 
 \toprule[1.5pt]
                 \multicolumn{1}{c|}{} & \multicolumn{5}{c|}{\textbf{Test Accuracy (\%)}}                              & \multicolumn{5}{c}{\textbf{Factor Speedup}}                                   \\ \midrule
 \textbf{Method}  & \textbf{2.9\%} & \textbf{5.7\%} & \textbf{8.6\%} & \textbf{11.4\%} & \textbf{14.3\%} & \textbf{2.9\%} & \textbf{5.7\%} & \textbf{8.6\%} & \textbf{11.4\%} & \textbf{14.3\%} \\ \midrule 
\textcolor{blue}{CAL-ER}  & 73.2 {\tiny $\pm$ 1.7} &	75.4 {\tiny $\pm$ 0.8}&	76.4 {\tiny $\pm$ 1.0}&	77.0 {\tiny $\pm$ 2.0}&	77.3 {\tiny $\pm$ 1.4}& 1.7x&	2.8x&	3.9x&	5.0x&	6.1x \\
\textcolor{blue}{CAL-MIR} & \textbf{75.1} {\tiny $\pm$ 0.2}&	75.5 {\tiny $\pm$ 1.2}&	76.6 {\tiny $\pm$ 1.0}&	76.5 {\tiny $\pm$ 0.4}&	77.0 {\tiny $\pm$ 0.3}& 0.8x&	1.2x&	1.6x&	2.0x&	2.4x\\
\textcolor{blue}{CAL-DER} & 71.5 {\tiny $\pm$ 2.7}&	74.9 {\tiny $\pm$ 3.2}&	75.7 {\tiny $\pm$ 1.5}&	76.9 {\tiny $\pm$ 1.6}&	78.3 {\tiny $\pm$ 0.8}& 1.7x&	2.7x&	3.7x&	4.8x&	5.8x \\
\textcolor{blue}{CAL-SD} & 73.6 {\tiny $\pm$ 1.9}&	74.9 {\tiny $\pm$ 1.1}&	\textbf{77.7} {\tiny $\pm$ 1.3}&	76.4 {\tiny $\pm$ 0.3}&	78.0 {\tiny $\pm$ 0.9}& 1.7x&	2.7x&	3.7x&	4.8x&	5.8x \\
\textcolor{blue}{CAL-SDS2} & 74.7 {\tiny $\pm$ 2.8}&  75.5 {\tiny $\pm$ 1.0}&         77.2  {\tiny $\pm$ 0.9}&  \textbf{78.1} {\tiny $\pm$ 0.8}& \textbf{79.2} {\tiny $\pm$ 0.5}& 1.4x&   2.1x&   2.9x&   3.7x&   4.5x \\
\midrule
AL w/ WS & 74.6 {\tiny $\pm$ 0.7}&	\textbf{76.1} {\tiny $\pm$ 0.4}&	76.3 {\tiny $\pm$ 1.0}&	76.3 {\tiny $\pm$ 1.5}&	77.2  {\tiny $\pm$ 0.9}& 1.2x&	1.7x&	1.6x&	1.8x&	1.8x \\
AL & 73.9 {\tiny $\pm$ 2.9}&	75.5 {\tiny $\pm$ 0.5}&	76.6 {\tiny $\pm$ 2.0}&	76.3 {\tiny $\pm$ 0.9}&	77.3  {\tiny $\pm$ 1.6}& 1.0x           & 1.0x & 1.0x & 1.0x & 1.0x\\ \midrule

\end{tabular}
}
\caption{COLA Results.}
\label{tab:cola_uncert}
\end{table}

\begin{figure}[H]
    \centering
    \includegraphics[width=1.0\linewidth]{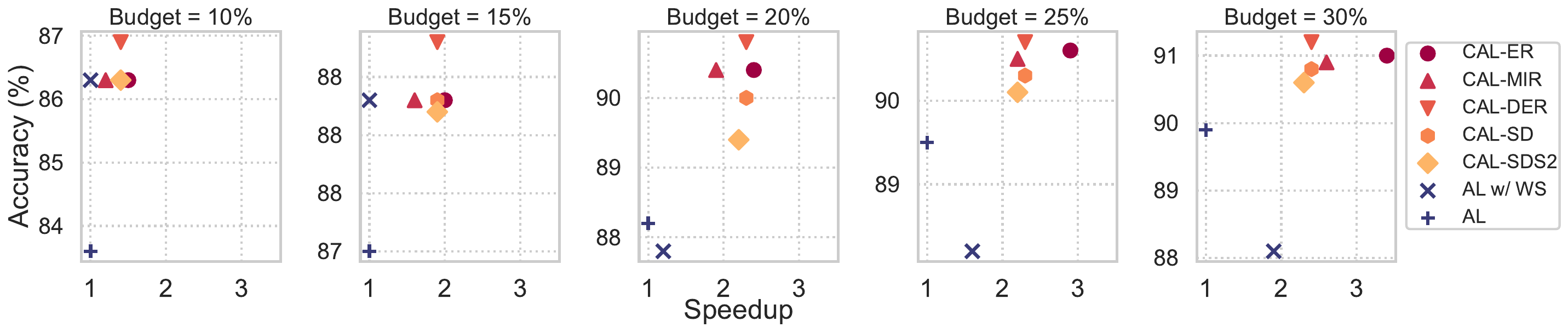} 
    \caption{Single-Cell Cell-Type Identity Classification Results}
    \label{fig:singlecell}
\end{figure}

\paragraph{Single-Cell Cell-Type Identity} Please refer to table~\ref{tab:cell_uncert} and figure~\ref{fig:singlecell}.

\begin{table}[H]
\centering
\scalebox{0.85}{
\begin{tabular}{c|ccccc|ccccc} 
 \toprule[1.5pt]
                 \multicolumn{1}{c|}{} & \multicolumn{5}{c|}{\textbf{Test Accuracy (\%)}}                              & \multicolumn{5}{c}{\textbf{Factor Speedup}}                                   \\ \midrule
 \textbf{Method}  & \textbf{10\%} & \textbf{15\%} & \textbf{20\%} & \textbf{25\%} & \textbf{30\%} & \textbf{10\%} & \textbf{15\%} & \textbf{20\%} & \textbf{25\%} & \textbf{30\%} \\ \midrule 
 \textcolor{blue}{CAL-ER} & 86.3 {\tiny $\pm$ 0.1} & 88.3 {\tiny $\pm$ 0.1} & 89.7 {\tiny $\pm$ 0.3} & 90.6 {\tiny $\pm$ 0.2} & 91.0 {\tiny $\pm$ 0.1} & 1.5$\times$ &	2.0$\times$ &	2.4$\times$ & 2.9$\times$ &	3.4$\times$    \\

\textcolor{blue}{CAL-MIR} & 86.3 {\tiny $\pm$ 0.1} & 88.3 {\tiny $\pm$ 0.1} & 89.7 {\tiny $\pm$ 0.2} & 90.5 {\tiny $\pm$ 0.2} & 90.9 {\tiny $\pm$ 0.2} & 1.2$\times$ &	1.6$\times$ &	1.9$\times$ &	2.2$\times$ & 2.6$\times$ \\

\textcolor{blue}{CAL-DER} & \textbf{86.9} {\tiny $\pm$ 0.3} & \textbf{88.8} {\tiny $\pm$ 0.3} & \textbf{89.9} {\tiny $\pm$ 0.3} & \textbf{90.7} {\tiny $\pm$ 0.2} & \textbf{91.2} {\tiny $\pm$ 0.1} & 1.4$\times$ &	1.9$\times$ &	2.3$\times$ &	2.8$\times$ &	3.3$\times$\\

\textcolor{blue}{CAL-SD} & 86.3 {\tiny $\pm$ 0.1} & 88.3 {\tiny $\pm$ 0.1} & 89.5 {\tiny $\pm$ 0.2} & 90.3 {\tiny $\pm$ 0.2} & 90.8 {\tiny $\pm$ 0.2} & 1.4$\times$ &	1.9$\times$ &	2.3$\times$ &	2.8$\times$ &	3.3$\times$      \\

\textcolor{blue}{CAL-SDS2} & 86.3 {\tiny $\pm$ 0.1}&	88.2{\tiny $\pm$ 0.1}&	89.2 {\tiny $\pm$ 0.3}&	90.1{\tiny $\pm$ 0.2} &	90.6{\tiny $\pm$ 0.1} & 1.4$\times$ &	1.9$\times$ &	2.3$\times$ &	2.8$\times$ &	3.3$\times$ \\
\midrule
AL w/ WS & 86.3	{\tiny $\pm$ 0.1} & 88.3 {\tiny $\pm$ 0.1} &	88.4 {\tiny $\pm$ 0.8} & 88.2 {\tiny $\pm$ 0.8}&	88.1 {\tiny $\pm$ 0.8}& 1.0$\times$ &	1.0$\times$ &	1.2$\times$ &	1.6$\times$ &	1.9$\times$ \\
AL & 83.6 {\tiny $\pm$ 1.0} & 87.0  {\tiny $\pm$ 0.3}& 88.6 {\tiny $\pm$ 0.1}&	89.5 {\tiny $\pm$ 0.2}& 89.9 {\tiny $\pm$ 0.3} & 1.0$\times$ & 1.0$\times$ & 1.0$\times$ & 1.0$\times$ & 1.0$\times$ \\ \midrule

\end{tabular}
  }
  \caption{Single-Cell Cell-Type Identity Classification Results}
  \label{tab:cell_uncert}
\end{table}

\subsection{Out-of-distribution (OOD) generalization}
\label{sec:appen_OOD}
Here we show results from the figure~\ref{fig:robustness} in a tabular form in table~\ref{tab: robustnesss}. We compare the performance of CAL methods with baseline AL for OOD generalization. Reported mean values and standard deviations are computed over three different random seeds. As shown in the figure~\ref{fig:robustness} in the main paper, and the table below, CAL methods are as robust to perturbations as baseline AL, despite the speedup. Finally, in table~\ref{tab:robustnesss_difference} we report that CAL-SDS2 is better compared to other CAL methods, on average. 


\begin{table}[H]
\centering
\scalebox{0.85}{\begin{tabular}{lccccc|cc}
\textbf{Corruption}              & \textcolor{blue}{CAL-ER} & \textcolor{blue}{CAL-MIR} & \textcolor{blue}{CAL-DER} & \textcolor{blue}{CAL-SD} & \textcolor{blue}{CAL-SDS2} & AL   & AL w/ WS \\
\toprule[1.5pt]
saturate           & 90.4 \tiny $\pm$ 0.4   & 90.7 \tiny $\pm$ 0.2   & 90.4 \tiny $\pm$ 0.1   & 90.5 \tiny $\pm$ 0.4   & 90.8 \tiny $\pm$ 0.2    & 90.7 \tiny $\pm$ 0.4 & 90.8 \tiny $\pm$ 0.2   \\
impulse noise     & 52.4 \tiny $\pm$ 1.4  & 53.4  \tiny $\pm$ 0.8  & 53.3 \tiny $\pm$ 3.9   & 53.6 \tiny $\pm$ 2.6  & 55.8 \tiny $\pm$ 1.3     & 54.5 \tiny $\pm$ 2.9 & 48.7 \tiny $\pm$ 2.7    \\
defocus blur      & 79.6 \tiny $\pm$ 1.3  & 81.6  \tiny $\pm$ 0.4  & 79.3 \tiny $\pm$ 1.2   & 80.6 \tiny $\pm$ 1.1  & 79.6 \tiny $\pm$ 1.7     & 80.9 \tiny $\pm$ 1.0 & 80.4 \tiny $\pm$ 0.7    \\
contrast           & 74.6 \tiny $\pm$ 0.3  & 76.1 \tiny $\pm$ 0.8   & 73.8 \tiny $\pm$ 0.9   & 75.1 \tiny $\pm$ 1.0  & 74.2 \tiny $\pm$ 0.6     & 77.1 \tiny $\pm$ 1.4 & 76.4 \tiny $\pm$ 2.5    \\
frost              & 76.0 \tiny $\pm$ 0.4  & 76.1 \tiny $\pm$ 2.0   & 73.8 \tiny $\pm$ 1.0   & 75.4 \tiny $\pm$ 1.5  & 76.9 \tiny $\pm$ 1.2     & 75.4 \tiny $\pm$ 1.6 & 77.0 \tiny $\pm$ 0.3    \\
speckle noise     & 61.7 \tiny $\pm$ 1.7  & 61.7 \tiny $\pm$ 2.6   & 60.9 \tiny $\pm$ 3.7   & 61.6 \tiny $\pm$ 3.8  & 64.3 \tiny $\pm$ 0.5     & 61.7 \tiny $\pm$ 0.3 & 59.4 \tiny $\pm$ 1.0    \\
pixelate           & 74.4 \tiny $\pm$ 1.3   & 73.9 \tiny $\pm$ 2.2   & 75.2 \tiny $\pm$ 0.9   & 74.8 \tiny $\pm$ 1.6  & 76.5 \tiny $\pm$ 0.7     & 75.9 \tiny $\pm$ 1.3 & 77.0 \tiny $\pm$ 1.3    \\
zoom blur         & 74.4 \tiny $\pm$ 1.0  & 76.7 \tiny $\pm$ 1.0   & 74.4 \tiny $\pm$ 2.3   & 75.4 \tiny $\pm$ 1.5  & 74.1 \tiny $\pm$ 2.8     & 75.9 \tiny $\pm$ 1.9 & 75.3 \tiny $\pm$ 1.0    \\
elastic transform & 82.5 \tiny $\pm$ 0.1  & 83.5 \tiny $\pm$ 0.2   & 82.0 \tiny $\pm$ 0.5   & 82.1 \tiny $\pm$ 0.8  & 82.6 \tiny $\pm$ 0.6     & 82.0 \tiny $\pm$ 1.3 & 82.9 \tiny $\pm$ 0.2    \\
spatter            & 81.8 \tiny $\pm$ 0.7  & 82.9 \tiny $\pm$ 1.2   & 82.4 \tiny $\pm$ 0.6   & 82.6 \tiny $\pm$ 0.5  & 82.8 \tiny $\pm$ 0.8     & 83.0 \tiny $\pm$ 1.0 & 82.5 \tiny $\pm$ 1.2    \\
snow               & 80.3 \tiny $\pm$ 0.4  & 80.2 \tiny $\pm$ 0.7   & 79.5 \tiny $\pm$ 0.8   & 79.7 \tiny $\pm$ 0.7  & 80.8 \tiny $\pm$ 0.1     & 80.1 \tiny $\pm$ 0.8 & 80.6 \tiny $\pm$ 0.3    \\
fog                & 86.8 \tiny $\pm$ 0.3  & 86.9 \tiny $\pm$ 0.3   & 85.9 \tiny $\pm$ 0.9   & 86.3 \tiny $\pm$ 0.5  & 86.5 \tiny $\pm$ 0.4     & 86.6 \tiny $\pm$ 0.3 & 87.9 \tiny $\pm$ 0.7    \\
Gaussian noise    & 46.6 \tiny $\pm$ 1.0  & 46.6 \tiny $\pm$ 3.3   & 44.9 \tiny $\pm$ 6.3   & 46.7 \tiny $\pm$ 4.7  & 50.7 \tiny $\pm$ 0.9     & 45.9 \tiny $\pm$ 0.4 & 43.2 \tiny $\pm$ 1.1    \\
brightness         & 92.1 \tiny $\pm$ 0.3   & 92.4 \tiny $\pm$ 0.3   & 92.1 \tiny $\pm$ 0.3   & 92.2 \tiny $\pm$ 0.1  & 92.5 \tiny $\pm$ 0.1     & 92.7 \tiny $\pm$ 0.1 & 92.6 \tiny $\pm$ 0.2    \\
Gaussian blur     & 69.7 \tiny $\pm$ 2.2  & 72.3 \tiny $\pm$ 1.3   & 69.3 \tiny $\pm$ 2.5   & 71.3 \tiny $\pm$ 1.5  & 69.3 \tiny $\pm$ 2.8     & 71.6 \tiny $\pm$ 1.6 & 70.4 \tiny $\pm$ 1.3    \\
motion blur       & 75.1 \tiny $\pm$ 1.2  & 77.1 \tiny $\pm$ 1.2   & 74.4 \tiny $\pm$ 0.9   & 75.1 \tiny $\pm$ 1.6  & 74.5 \tiny $\pm$ 0.7     & 74.7 \tiny $\pm$ 0.6 & 76.8 \tiny $\pm$ 0.4    \\
shot noise        & 58.6 \tiny $\pm$ 1.5  & 58.5 \tiny $\pm$ 2.9   & 57.3 \tiny $\pm$ 4.8   & 58.6 \tiny $\pm$ 4.1  & 61.6 \tiny $\pm$ 0.5     & 58.3 \tiny $\pm$ 0.2 & 56.1 \tiny $\pm$ 1.0    \\
jpeg compression  & 79.0 \tiny $\pm$ 1.5  & 78.4 \tiny $\pm$ 0.5   & 78.5 \tiny $\pm$ 0.2   & 78.6 \tiny $\pm$ 0.5  & 79.1 \tiny $\pm$ 0.1     & 77.7 \tiny $\pm$ 0.1 & 77.7 \tiny $\pm$ 0.1    \\
glass blur        & 51.8 \tiny $\pm$ 1.4  & 54.6 \tiny $\pm$ 3.0   & 48.6 \tiny $\pm$ 2.8   & 50.7 \tiny $\pm$ 2.1  & 53.9 \tiny $\pm$ 2.4     & 49.4 \tiny $\pm$ 3.1 & 52.6 \tiny $\pm$ 2.1 \\ \midrule
\end{tabular}}
  \caption{Accuracy (in \%) comparison of CAL methods with the baseline on the CIFAR-10C dataset. Results were reported as an average over three random seeds. Models trained with CAL procedure perform statistically similar to the one trained with baseline AL.}

  \label{tab: robustnesss}
\end{table}

\begin{table}[H]
\centering
\scalebox{0.85}{\begin{tabular}{lccccc}
\textbf{Corruption}         & CAL-ER/AL       & CAL-MIR/AL      & CAL-DER/AL      & CAL-SD/AL       & CAL-SDS2/AL     \\
\toprule[1.5pt]
saturate           & 0.3957          & 0.9104          & 0.2118          & 0.4516          & 0.7961          \\
impulse noise     & 0.3209          & 0.5692          & 0.7029          & 0.7134          & 0.5374          \\
defocus blur      & 0.2477          & 0.2794          & 0.1561          & 0.7854          & 0.3178          \\
contrast           & \textbf{0.0335 (+)} & 0.3023          & \textbf{0.0256 (-)} & 0.1020          & \textbf{0.0250 (-)} \\
frost              & 0.5793          & 0.6639          & 0.2147          & 0.9548          & 0.2778          \\
speckle noise     & 0.9873          & 0.9922          & 0.7409          & 0.9739          & \textbf{0.0017 (+)} \\
pixelate           & 0.2243          & 0.2521          & 0.4661          & 0.3913          & 0.5636          \\
zoom blur         & 0.2824          & 0.5567          & 0.4279          & 0.7238          & 0.3910          \\
elastic transform & 0.5482          & 0.1295          & 0.9975          & 0.9154          & 0.5176          \\
spatter            & 0.1504          & 0.9401          & 0.4310          & 0.5843          & 0.7800          \\
snow               & 0.7823          & 0.8905          & 0.3747          & 0.5455          & 0.1976          \\
fog                & 0.4715          & 0.2534          & 0.2781          & 0.4512          & 0.7142          \\
gaussian noise    & 0.3514          & 0.7485          & 0.7836          & 0.8056          & \textbf{0.0013 (+)} \\
brightness         & \textbf{0.0309 (-)}          & 0.2042          & \textbf{0.0341 (-)} & \textbf{0.0091 (-)} & \textbf{0.0474 (-)} \\
gaussian blur     & 0.2969          & 0.5867          & 0.2605          & 0.8658          & 0.2997          \\
motion blur       & 0.6474          & \textbf{0.0368 (+)} & 0.6263          & 0.6801          & 0.7135          \\
shot noise        & 0.7252          & 0.8920          & 0.7311          & 0.9091          & \textbf{0.0004 (+)} \\
jpeg compression  & \textbf{0.0110 (+)}          & 0.1080          & \textbf{0.0016 (+)} & \textbf{0.0286 (+)} & \textbf{0.0000 (+)} \\
glass blur        & 0.2820          & 0.1047          & 0.7633          & 0.5807          & 0.1182         \\ \midrule    
\end{tabular}}
  \caption{Pairwise p-values were computed to compare each Confidence-Aware Learning (CAL) method with the standard Active Learning (AL) approach across various corruption tests. p-values less than 0.05 are highlighted in \textbf{bold}, and instances, where CAL methods outperform (+) or underperform (-) compared to standard AL, are indicated. Overall, our findings suggest that the disparities between CAL and AL methods are typically not statistically significant, implying that CAL generally does not compromise robustness. When a statistically significant difference does arise, CAL (particularly CAL-SDS2) tends to outperform AL more frequently than it falls short.}

  \label{tab:robustness_significance}

\end{table}

\begin{table}[htp]
\centering
\begin{tabular}{@{}cc@{}}
\textbf{Method}   & \textbf{Average Accuracy difference (in \%)} \\
\toprule[1.5pt]
\textcolor{blue}{CAL-ER}   & -0.34                               \\
\textcolor{blue}{CAL-MIR}  & 0.49                                \\
\textcolor{blue}{CAL-DER}  & -0.96                               \\
\textcolor{blue}{CAL-SD}   & -0.17                               \\
\textcolor{blue}{CAL-SDS2} & 0.63                                \\ \midrule
AL w/ WS & -0.32                               \\
\midrule
\end{tabular}
  \caption{The average difference in the accuracy (in \%) of models trained with baseline AL from models trained with CAL, on different benchmarks across the CIFAR10-C dataset. Higher numbers are better. We can see that CAL-SDS2 is better compared to other CAL methods, on average. We hypothesize that this can be attributed to submodular sampling from history.}
\label{tab:robustnesss_difference}
\end{table}

\subsection{Correlation between CAL and baseline AL}
\begin{figure}[H]
    \centering
    \includegraphics[width=0.85\linewidth]{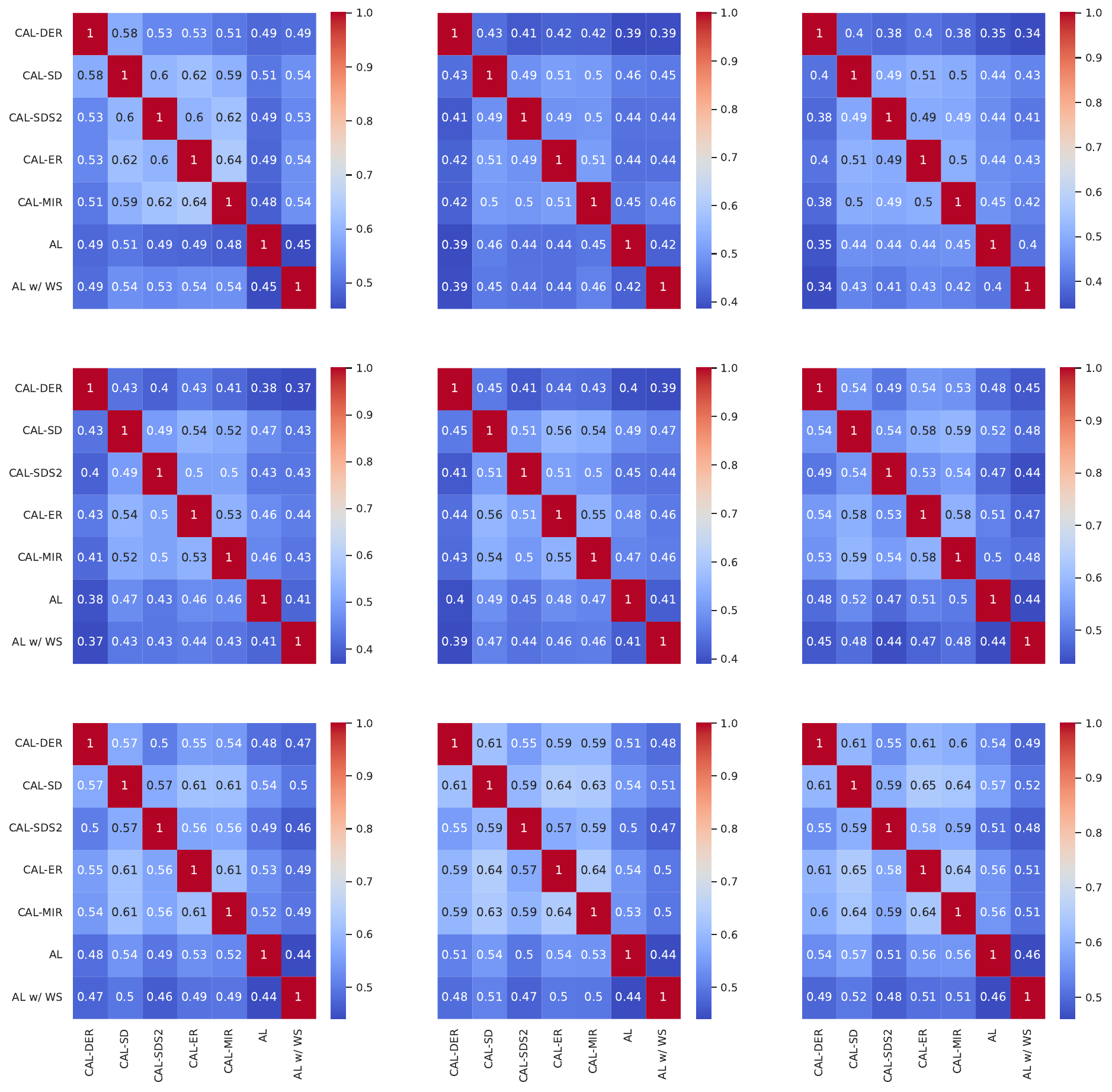} 
    \caption{Pearson correlation between uncertainty scores of models on held out set after every query round. We are showing nine query rounds in row-major order (so the top left is after the first query round, the top middle is after the second, and so on). Positive correlation of uncertainty scores suggests that the nature of examples the models are uncertain about, and thus likely to be chosen at every query round, is similar between the CAL-trained models and baseline AL-trained models.}
    \label{fig:all_correlation}
\end{figure}

In figure~\ref{fig:all_correlation}, we provide the Pearson correlation between the uncertainty scores of models on the held-out test set, before every query round. A positive correlation between CAL models and baseline AL models before every query round suggests that the nature of examples chosen by the CAL models is similar to that of baseline AL models. Note that each entry in the correlation matrix is averaged over three random seeds corresponding to the random initialization of each model. 

\subsection{Hyperparameters}
\label{sec:appen_hyperparameters}
For every dataset and every CAL/AL strategy, learning rate ($lr$) and batch size ($\numnew$) are chosen based on whichever setting achieves the highest performance on standard AL. This means CAL methods can still improve if we tune either of learning rate or batch size. Replay size is critical for the performance of continual learning algorithms. On the one hand, we do not want the replay batch size, $\numfromhist$, to be too small since then we will forget some history. But on the other hand, we also do not want $\numfromhist$ to be too large since there will be a computational cost associated with that. Considering the mentioned constraints, for all CAL methods, we, therefore, set replay size as $\numfromhist \in \{\numnew, 2\numnew\}$ (used in all CAL methods). Via experimentation on a subset of the datasets, we found $\numfromhist$ less than or greater than this range suffered either from forgetting (when too small) or extra computation without an accuracy benefit (when too large).  We set $\alpha \in \{0.1, 0.25, 0.5, 0.75\}$ (used in CAL-DER, CAL-SD, and CAL-SDS2), $\beta \in \{0.75, 1\}$ (used in CAL-DER). The scale and size of the search space for $\alpha$ and $\beta$ is inspired from~\citet{DER2020}. Lastly, $\sigma \in \{0.1, 1\}$ (used in CAL-SDS2), and $\lambda \in \{0.1, 1, 10\}$ (used in CAL-SDS2).

We select the configuration for each CAL model that achieves the highest accuracy. $``c"$ is the hyperparameter used in CAL-MIR and CAL-SDS2 to subsample the history before finding the $\numfromhist$ samples to replay, but this parameter is not tuned for any of the presented results. We list the specific set of hyperparameters we use for all the main experimental results in this section. 
\subsubsection{FMNIST} 
All experiments for FMNIST used a ResNet-18 with an SGD optimizer, with learning rate of 0.01 and batch size of 64. For all the CAL methods, we fix $\numfromhist=128$. A NVIDIA GeForce RTX 1080 GPU was used to run all the reported experiments.


\paragraph{CAL-MIR} $c = 256$

\paragraph{CAL-DER} $\alpha = 0.1$, $\beta = 1$

\paragraph{CAL-SD}  $\alpha = 0.25$ 

\paragraph{CAL-SDS2} $c=256$, $\alpha = 0.25$, $\sigma = 0.1$, $\lambda = 1$

\subsubsection{CIFAR-10} 
All experiments for CIFAR-10 used a ResNet-18 with an SGD optimizer, with learning rate of 0.02 and a batch size of 20. For all the CAL methods, we fix $\numfromhist=40$. Training is done on an NVIDIA GeForce RTX 2080.


\paragraph{CAL-MIR} $c = 100$

\paragraph{CAL-DER} $\alpha = 0.1$, $\beta = 1$

\paragraph{CAL-SD} $\alpha = 0.25$ 

\paragraph{CAL-SDS2} $c = 100$, $\alpha = 0.25$, $\sigma = 0.1$, $\lambda = 0.1$

\subsubsection{MedMNIST} 
All experiments for MedMNIST used a ResNet-18 with an Adam optimizer, with learning rate of 0.001 and a batch size of 128. For all CAL methods, we fix $\numfromhist = 128$. All reported models were trained on an NVIDIA GeForce RTX 2080.

\paragraph{CAL-MIR} $c = 270$

\paragraph{CAL-DER} $\alpha = 0.1$, $\beta = 1$

\paragraph{CAL-SD} $\alpha = 0.5$ 

\paragraph{CAL-SDS2} $c = 270$, $\alpha = 0.5$, $\sigma = 0.1$, $\lambda = 10$

\subsubsection{Amazon Polarity Review} 
Throughout our experiments, we sample 2M sentences and use them as the total training set instead. We use Adam optimizer with default parameters with a learning rate of 0.001 and a batch size of 128 for 6 epochs. For all the CAL methods, we fix $\numfromhist=128$. All reported models were trained on an NVIDIA GeForce 1080 Ti.
\paragraph{CAL-MIR} $c = 256$,

\paragraph{CAL-DER} $\alpha = 0.25$, $\beta = 0.75$

\paragraph{CAL-SD} $\alpha = 0.5$ 

\paragraph{CAL-SDS2} $c = 256$, $\alpha = 0.75$, $\sigma = 1$, $\lambda = 1$

\subsubsection{COLA } For all of our experiments we use Huggingface's transformer library \cite{wolf-etal-2020-transformers} and use a maximum sentence length of 100. We use Adam optimizer and a learning rate of $5\cdot 10^{-5}$, use a batch size of 25 and $\numfromhist=25$. Models were trained on a single NVIDIA GeForce 1080 Ti.

\paragraph{CAL-MIR} $c=50$
\paragraph{CAL-DER} $\alpha=0.25$, $\beta=0.75$
\paragraph{CAL-SD} $\alpha=0.75$
\paragraph{CAL-SDS2} $c=50$, $\alpha=0.5$, $\sigma=1$, $\lambda=1$.

\subsubsection{Single-Cell Cell-Type Identity Classification} 
 All experiments use SGD optimizer with standard parameters with learning rate of 0.001 and a batch size 128. For all the CAL methods, we fix $\numfromhist=128$. Training is done on an NVIDIA A100-PCIE-40GB. 
\paragraph{CAL-MIR} $c = 200$,

\paragraph{CAL-DER} $\alpha = 0.1$, $\beta = 1$

\paragraph{CAL-SD} $\alpha = 1$ 

\paragraph{CAL-SDS2} $c = 100$, $\alpha = 0.25$, $\sigma = 0.1$, $\lambda = 1$

\subsection{Hyperparameter Sensitivity Analysis for Proposed Methods}
\label{sec:appen_sensitivity}
In this section, we provide a hyperparameter sensitivity analysis methods for CAL-SD and CAL-SDS2 on CIFAR-10. For CAL-SD, only $\alpha$ is tuned and the best value is used for CAL-SDS2. Therefore, we tuned 1 hyperparameter for CAL-SD while $\sigma$ and $\lambda$ are tuned for CAL-SDS2. Note that the differences in final test accuracy at 50\% budget on CIFAR-10 across different configurations are negligible, as shown in the tables below. 

\begin{table}[htp]
\centering
\begin{tabular}{|c|c|c|c|c|}
\hline
$\alpha$ & 0.1     & 0.25    & 0.75    & 0.9 \\ \hline
      & 93.79 & 93.90 & 93.58 & 93.38 \\ \hline
\end{tabular}
  \caption{CAL-SD Sensitivity}
\end{table}

\begin{table}[htp]
\centering
\begin{tabular}{|c|c|c|c|c|c|}
\hline
\diagbox[]{$\sigma$}{$\lambda$} & 0.1     & 0.5       & 1       & 5       & 10      \\ \hline
0.1    & 94.20 & 94.27   & 94.32 & 94.34 & 94.16 \\ \hline
1      & 94.33 & 94.22   & 94.28 & 94.15 & \textbf{94.44} \\ \hline
10     & 94.32 & 94.29 & 94.27 & 94.30 & 94.18 \\ \hline
\end{tabular}
\caption{CAL-SDS2 Sensitivity}
\end{table}

\section{A primer on Continual Learning}
\label{sec:appen_CL}
We define $\mathcal{D}_{1:n} = \bigcup_{i \in [n]} \mathcal{D}_i$. In CL, the  dataset consists of $T$ tasks $\{\mathcal{D}_1,...,\mathcal{D}_T\}$ that are presented to the model sequentially, where $\mathcal{D}_t = \{(x_i, y_i)\}_{i \in N_t}$, $N_t$ are the task-$t$ sample indices, and $n_t=|N_t|$. At time $t \in [T]$, the data/label pairs are sampled from the current task $(x,y) \sim \mathcal{D}_t$, and the model has only limited access to the history $\mathcal{D}_{1:t-1}$.
If the model is trained only on $\mathcal{D}_t$ using standard optimization algorithms, the model will exhibit catastrophic forgetting. The CL objective is to efficiently adapt the model to $\mathcal{D}_t$ while retaining the performance on the history. 
Given a loss function $\ell: \mathcal{X} \times \mathcal{Y} \mapsto \mathbb{R}$, initial parameters $\theta_{t-1}$, and a model $f$, $\theta_t$ can be obtained as the solution to the CL optimization problem \citep{aljundi2019gradient, AGEM2019, GEM2017}:
{
\begin{alignat*}{2}
 &\min_{\theta}   \mathop{\mathbb{E}}_{(x,y) \sim \mathcal{D}_t}  \ell(y, f(x; \theta))  \\
\text{s.t.}  &\hspace{-1ex} \mathop{\mathbb{E}}_{(x',y') \sim \mathcal{D}_{1:t-1}} \ell \Biggl(y', f (x'; \theta))\Biggr) & \leq \hspace{-1ex} \mathop{\mathbb{E}}_{(x',y') \sim \mathcal{D}_{1:t-1}} \hspace{-1ex} \ell\Biggl(y', f (x'; \theta_{t-1}))\Biggr)
\end{alignat*}}%

\section{Results for Additional Active Learning Strategies}
\label{sec:appen_other_acquisition_func}
In this section, we demonstrate that CAL methods can accelerate AL strategies other than entropy sampling without incurring any significant performance drops. We test multiple AL strategies on FMNIST \cite{fmnist} and CIFAR-10 \cite{cifar10}. Note that the speedups are approximately the same as the ones reported in Section A since the training time is generally independent of the selected AL strategy.

\subsection{Overview of Strategies}
\paragraph{Margin Score Sampling} This strategy is another form of uncertainty sampling \cite{settles2009active} as described in the main paper. Instead of the entropy of $f(x; \theta)$, the margin score is used as the entropy score i.e., $h(x) \triangleq 1 - (f(x;\theta)_i - f(x; \theta)_j)$ where $i$ and $j$ are the indices corresponding to the highest and second highest values of $f(x; \theta)$ respectively.

\paragraph{FASS} FASS \cite{fass2015} is a two-staged selection method that uses both uncertainty sampling and submodular maximization. Initially, a set of samples $\mathcal{A}$ of cardinality $c * b_t$ is chosen from $\mathcal{U}$ using uncertainty sampling, where $c > 1$ is a tuneable hyperparameter. Next, $U_t$ is constructed by greedily selecting samples that maximize a submodular set function $G: 2^{\mathcal{A}} \rightarrow \mathbb{R}_{+}$ defined on a ground set $\mathcal{A}$. Entropy is once again used as the uncertainty metric for the initial stage. For the second stage, $G$ is defined to be the facility location function \cite{fass2015} expressed below: 

\begin{equation}
    G(\mathcal{S}) = \sum_{x_i \in \mathcal{A}} \max_{x_j \in \mathcal{S}} w_{ij},
\end{equation}

where $\mathcal{S} \subseteq \mathcal{A}$ and $w_{ij}$ is a similarity score between samples $x_i$ and $x_j$. In our experiments, $w_{ij} = \exp{(-\|z_i - z_j \|^2 / 2 \sigma^2)}$ where $z_i$ is the penultimate layer representation of model $f$ for $x_i$ and $\sigma$ is a hyperparameter.

\paragraph{GLISTER} GLISTER \cite{glister2021} solves a bi-level optimization problem in order to select samples to label. Specifically, GLISTER solves 

\begin{equation}
    \argmax_{\mathcal{S} \subseteq \mathcal{U}_t, |\mathcal{S}| \leq b_t} LL_V(\argmax_{\theta} LL_T(\theta, \mathcal{S}), \mathcal{V})
\end{equation}

where $LL_V$ is the log-likelihood on the validation set $\mathcal{V}$, and $LL_T$ is the log-likelihood on the subset $\mathcal{S}$.

\subsection{Results}

\begin{table}[H]
\centering
\scalebox{1}{
\begin{tabular}{c|ccccc} 
 \toprule[1.5pt]
\multicolumn{1}{c|}{} &\multicolumn{5}{c}{\textbf{Test Accuracy (\%)}}\\ \midrule
\textbf{Method}  & \textbf{10\%} & \textbf{15\%} & \textbf{20\%} & \textbf{25\%} & \textbf{30\%}  \\ \midrule 

\textcolor{blue}{CAL-ER} & \textbf{92.8} {\tiny $\pm$ 0.1} &	\textbf{94.1} {\tiny $\pm$ 0.1} &	94.8 {\tiny $\pm$ 0.1} &	\textbf{95.1} {\tiny $\pm$ 0.3} &	\textbf{95.2} {\tiny $\pm$ 0.2} \\

\textcolor{blue}{CAL-MIR}&  92.6 {\tiny $\pm$ 0.2} &	\textbf{94.1} {\tiny $\pm$ 0.4} &	\textbf{94.9} {\tiny $\pm$ 0.2} &	95.0 {\tiny $\pm$ 0.2} &	\textbf{95.2} {\tiny $\pm$ 0.2} \\

\textcolor{blue}{CAL-DER}&  91.8 {\tiny $\pm$ 0.5} &	93.1 {\tiny $\pm$ 0.1} &	94.3 {\tiny $\pm$ 0.3} &	94.6 {\tiny $\pm$ 0.1} &	94.8 {\tiny $\pm$ 0.2} \\

\textcolor{blue}{CAL-SD} &  92.5 {\tiny $\pm$ 0.1} &	93.8 {\tiny $\pm$ 0.1} &	94.8 {\tiny $\pm$ 0.0} &	\textbf{95.1} {\tiny $\pm$ 0.2} &	\textbf{95.2} {\tiny $\pm$ 0.0} \\

\textcolor{blue}{CAL-SDS2} & 87.8 {\tiny $\pm$ 1.1} &	93.4 {\tiny $\pm$ 0.1} &	94.6 {\tiny $\pm$ 0.1} &	95.0 {\tiny $\pm$ 0.2} &	\textbf{95.2} {\tiny $\pm$ 0.1} \\

\midrule

AL w/ WS & \textbf{92.8} {\tiny $\pm$ 0.0} &	94.0 {\tiny $\pm$ 0.3} &	94.6 {\tiny $\pm$ 0.1} &	94.8 {\tiny $\pm$ 0.1} &	95.0 {\tiny $\pm$ 0.2} \\

AL & 92.7 {\tiny $\pm$ 0.1} &	\textbf{94.1} {\tiny $\pm$ 0.3} &	\textbf{94.9} {\tiny $\pm$ 0.1} &	95.0 {\tiny $\pm$ 0.2} &	\textbf{95.2} {\tiny $\pm$ 0.1} \\ \midrule  
\end{tabular}}
  \caption{FMNIST with Margin Score Sampling}
\end{table}

\begin{table}[H]
\centering
\scalebox{1}{
\begin{tabular}{c|ccccc} 
 \toprule[1.5pt]
\multicolumn{1}{c|}{} &\multicolumn{5}{c}{\textbf{Test Accuracy (\%)}}\\ \midrule
\textbf{Method}  & \textbf{10\%} & \textbf{20\%} & \textbf{30\%} & \textbf{40\%} & \textbf{50\%}  \\ \midrule 

\textcolor{blue}{CAL-ER} & 81.5	{\tiny $\pm$ 0.1} &	89.3 {\tiny $\pm$ 0.1}	&	92.2{\tiny $\pm$ 0.2}	&	93.4{\tiny $\pm$ 0.1}	&	93.8 {\tiny $\pm$ 0.0}\\

\textcolor{blue}{CAL-MIR} & 81.9 {\tiny $\pm$ 0.1} & 89.6 {\tiny $\pm$ 0.2}	& 92.2 {\tiny $\pm$ 0.4} & 93.6 {\tiny $\pm$ 0.0} & 94.0 {\tiny $\pm$ 0.2}  \\

\textcolor{blue}{CAL-DER} & 83.0 {\tiny $\pm$ 0.2} & 89.5 {\tiny $\pm$ 0.2} & 92.2 {\tiny $\pm$ 0.2} & 93.2 {\tiny $\pm$ 0.2} & 93.6 {\tiny $\pm$ 0.0} \\

\textcolor{blue}{CAL-SD} & 82.6 {\tiny $\pm$ 0.4} & 89.9 {\tiny $\pm$ 0.4} & 92.4 {\tiny $\pm$ 0.2} & 93.5 {\tiny $\pm$ 0.1} & 93.8 {\tiny $\pm$ 0.2}\\

\textcolor{blue}{CAL-SDS2} & 82.5 {\tiny $\pm$ 0.2} & 90.2 {\tiny $\pm$ 0.2} & 92.5  {\tiny $\pm$ 0.2} & \textbf{93.8} {\tiny $\pm$ 0.2} & \textbf{94.1} {\tiny $\pm$ 0.1}  \\
\midrule

AL w/ WS & \textbf{83.1 }{\tiny $\pm$ 0.1} & \textbf{90.3} {\tiny $\pm$ 0.3} & \textbf{93.0} {\tiny $\pm$ 0.2} & 93.5 {\tiny $\pm$ 0.3} & 93.6 {\tiny $\pm$ 0.2}  \\

AL & 75.1	{\tiny $\pm$ 1.2} &	87.1 {\tiny $\pm$ 1.0}	&	90.2{\tiny $\pm$ 0.5}	&	92.0{\tiny $\pm$ 0.0}	&	92.8 {\tiny $\pm$ 0.5}  \\ \midrule

\end{tabular}}
  \caption{CIFAR-10 with Margin Score Sampling}
\end{table}

\begin{table}[H]
\centering
\scalebox{1}{
\begin{tabular}{c|ccccc} 
 \toprule[1.5pt]
\multicolumn{1}{c|}{} &\multicolumn{5}{c}{\textbf{Test Accuracy (\%)}}\\ \midrule
\textbf{Method}  & \textbf{10\%} & \textbf{15\%} & \textbf{20\%} & \textbf{25\%} & \textbf{30\%}  \\ \midrule 

\textcolor{blue}{CAL-ER} & 92.6 {\tiny $\pm$ 0.1} & \textbf{93.9} {\tiny $\pm$ 0.2} & 94.6 {\tiny $\pm$ 0.2} & \textbf{95.0} {\tiny $\pm$ 0.1} & 94.9 {\tiny $\pm$ 0.0} \\

\textcolor{blue}{CAL-MIR} & 92.5 {\tiny $\pm$ 0.1} & 93.8 {\tiny $\pm$ 0.3} & 94.6 {\tiny $\pm$ 0.1} & 94.8 {\tiny $\pm$ 0.1} & 94.9 {\tiny $\pm$ 0.2} \\

\textcolor{blue}{CAL-DER} & 92.7 {\tiny $\pm$ 0.1} & 93.8 {\tiny $\pm$ 0.1} & 94.5 {\tiny $\pm$ 0.1} & 94.7 {\tiny $\pm$ 0.1}  & \textbf{95.0} {\tiny $\pm$ 0.2} \\

\textcolor{blue}{CAL-SD} & \textbf{92.8} {\tiny $\pm$ 0.1} & \textbf{93.9} {\tiny $\pm$ 0.1} & \textbf{94.7} {\tiny $\pm$ 0.1} & 94.8 {\tiny $\pm$ 0.3} & 94.9 {\tiny $\pm$ 0.1}  \\

\textcolor{blue}{CAL-SDS2} & \textbf{92.8} {\tiny $\pm$ 0.0} & 93.8 {\tiny $\pm$ 0.2} & 94.5 {\tiny $\pm$ 0.1} & 94.8 {\tiny $\pm$ 0.2} & 94.9 {\tiny $\pm$ 0.1} \\ \midrule

AL w/ WS & 92.5 {\tiny $\pm$ 0.1} & 93.8 {\tiny $\pm$ 0.3} & 94.0 {\tiny $\pm$ 0.2} & 94.3 {\tiny $\pm$ 0.2} & 94.3 {\tiny $\pm$ 0.0} \\

AL & 92.7 {\tiny $\pm$ 0.4} & \textbf{93.9} {\tiny $\pm$ 0.1} & 94.5 {\tiny $\pm$ 0.1} & 94.7 {\tiny $\pm$ 0.3} & 94.8 {\tiny $\pm$ 0.1} \\ \midrule
\end{tabular}}
  \caption{FMNIST with FASS}
\end{table}

\begin{table}[H]
\centering
\scalebox{1}{
\begin{tabular}{c|ccccc} 
 \toprule[1.5pt]
\multicolumn{1}{c|}{} &\multicolumn{5}{c}{\textbf{Test Accuracy (\%)}}\\ \midrule
\textbf{Method}  & \textbf{10\%} & \textbf{20\%} & \textbf{30\%} & \textbf{40\%} & \textbf{50\%}  \\ \midrule 

\textcolor{blue}{CAL-ER} & 82.2	{\tiny $\pm$ 0.2} &	89.8 {\tiny $\pm$ 0.2}	&	92.5{\tiny $\pm$ 0.2}	&	93.4{\tiny $\pm$ 0.4}	&	93.7 {\tiny $\pm$ 0.2}\\

\textcolor{blue}{CAL-MIR} & 82.2 {\tiny $\pm$ 0.3} & 89.4 {\tiny $\pm$ 0.2}	& 92.3 {\tiny $\pm$ 0.1} & 93.4 {\tiny $\pm$ 0.0} & 93.5 {\tiny $\pm$ 0.1}  \\

\textcolor{blue}{CAL-DER} & \textbf{83.1} {\tiny $\pm$ 0.3} & 89.7 {\tiny $\pm$ 0.2} & 91.9 {\tiny $\pm$ 0.1} & 93.1 {\tiny $\pm$ 0.2} & 93.5 {\tiny $\pm$ 0.1} \\

\textcolor{blue}{CAL-SD} & 83.0 {\tiny $\pm$ 0.3} & 90.0 {\tiny $\pm$ 0.3} & 92.5 {\tiny $\pm$ 0.1} & 93.5 {\tiny $\pm$ 0.1} & \textbf{94.0} {\tiny $\pm$ 0.1}\\

\textcolor{blue}{CAL-SDS2} & 83.0 {\tiny $\pm$ 0.1} & 90.1 {\tiny $\pm$ 0.1} & 92.7  {\tiny $\pm$ 0.2} & 93.5 {\tiny $\pm$ 0.2} & \textbf{94.0} {\tiny $\pm$ 0.0}  \\
\midrule

AL w/ WS & 82.8 {\tiny $\pm$ 0.4} & \textbf{90.3} {\tiny $\pm$ 0.1} & \textbf{92.8} {\tiny $\pm$ 0.2} & \textbf{93.6} {\tiny $\pm$ 0.1} & 93.7 {\tiny $\pm$ 0.3}  \\

AL & 72.5	{\tiny $\pm$ 2.0} &	86.6 {\tiny $\pm$ 0.4}	&	90.1{\tiny $\pm$ 0.4}	&	91.7{\tiny $\pm$ 0.2}	&	92.9 {\tiny $\pm$ 0.2}  \\ \midrule

\end{tabular}}
  \caption{CIFAR-10 with FASS}
\end{table}

\begin{table}[H]
\centering
\scalebox{1}{
\begin{tabular}{c|ccccc} 
 \toprule[1.5pt]
\multicolumn{1}{c|}{} &\multicolumn{5}{c}{\textbf{Test Accuracy (\%)}}\\ \midrule
\textbf{Method}  & \textbf{10\%} & \textbf{15\%} & \textbf{20\%} & \textbf{25\%} & \textbf{30\%}  \\ \midrule 
\textcolor{blue}{CAL-ER} & 92.6 {\tiny $\pm$ 0.0} & \textbf{93.9} {\tiny $\pm$ 0.2} & 94.3 {\tiny $\pm$ 0.1} & \textbf{94.7} {\tiny $\pm$ 0.1} & 94.7 {\tiny $\pm$ 0.2} \\
\textcolor{blue}{CAL-MIR} & 92.5 {\tiny $\pm$ 0.0} & \textbf{93.9} {\tiny $\pm$ 0.4} & 94.3 {\tiny $\pm$ 0.2} & 94.4 {\tiny $\pm$ 0.2} & 94.6 {\tiny $\pm$ 0.1} \\
\textcolor{blue}{CAL-DER} & \textbf{92.7} {\tiny $\pm$ 0.1} & \textbf{93.9} {\tiny $\pm$ 0.2} & 94.3 {\tiny $\pm$ 0.3} & \textbf{94.7} {\tiny $\pm$ 0.2}  & \textbf{94.9} {\tiny $\pm$ 0.3} \\
\textcolor{blue}{CAL-SD} & 92.6 {\tiny $\pm$ 0.1} & 93.8 {\tiny $\pm$ 0.1} & \textbf{94.4} {\tiny $\pm$ 0.3} & 94.6 {\tiny $\pm$ 0.1} & 94.7 {\tiny $\pm$ 0.1}  \\
\textcolor{blue}{CAL-SDS2} & 92.6 {\tiny $\pm$ 0.1} & \textbf{93.9} {\tiny $\pm$ 0.2} & \textbf{94.4} {\tiny $\pm$ 0.2} & 94.6 {\tiny $\pm$ 0.3} & 94.7 {\tiny $\pm$ 0.2} \\ \midrule
AL w/ WS & 92.5 {\tiny $\pm$ 0.1} & 93.6 {\tiny $\pm$ 0.1} & 93.9 {\tiny $\pm$ 0.1} & 94.1 {\tiny $\pm$ 0.1} & 94.3 {\tiny $\pm$ 0.1} \\
AL & 92.5 {\tiny $\pm$ 0.2} & 93.8 {\tiny $\pm$ 0.1} & 94.2 {\tiny $\pm$ 0.1} & 94.6 {\tiny $\pm$ 0.2} & 94.7 {\tiny $\pm$ 0.2} \\ \midrule
\end{tabular}}
  \caption{FMNIST with GLISTER}
\end{table}

\begin{table}[H]
\centering
\scalebox{1}{
\begin{tabular}{c|ccccc} 
 \toprule[1.5pt]
\multicolumn{1}{c|}{} &\multicolumn{5}{c}{\textbf{Test Accuracy (\%)}}\\ \midrule
\textbf{Method}  & \textbf{10\%} & \textbf{20\%} & \textbf{30\%} & \textbf{40\%} & \textbf{50\%}  \\ \midrule 
\textcolor{blue}{CAL-ER} & 81.7 {\tiny $\pm$ 0.3} & 89.2 {\tiny $\pm$ 0.2} & 91.9 {\tiny $\pm$ 0.2} & 93.0 {\tiny $\pm$ 0.1} & 93.3 {\tiny $\pm$ 0.1} \\
\textcolor{blue}{CAL-MIR} & 81.6 {\tiny $\pm$ 0.3} & 89.3 {\tiny $\pm$ 0.4} & 91.7 {\tiny $\pm$ 0.2} & 92.9 {\tiny $\pm$ 0.1} & 93.5 {\tiny $\pm$ 0.2} \\
\textcolor{blue}{CAL-DER} & \textbf{82.8} {\tiny $\pm$ 0.4} & 89.5 {\tiny $\pm$ 0.4} & 91.7 {\tiny $\pm$ 0.4} & 92.8{\tiny $\pm$ 0.6}  & 93.1 {\tiny $\pm$ 0.2} \\
\textcolor{blue}{CAL-SD} & 82.5 {\tiny $\pm$ 0.3} & \textbf{89.6} {\tiny $\pm$ 0.2} & \textbf{92.1} {\tiny $\pm$ 0.2} & 93.1 {\tiny $\pm$ 0.2} & 93.8 {\tiny $\pm$ 0.1}  \\
\textcolor{blue}{CAL-SDS2} & 81.4 {\tiny $\pm$ 0.4} & 89.1 {\tiny $\pm$ 0.2} & \textbf{92.1} {\tiny $\pm$ 0.2} & \textbf{93.2} {\tiny $\pm$ 0.3} & \textbf{93.9} {\tiny $\pm$ 0.1} \\ \midrule
AL w/ WS & 81.7 {\tiny $\pm$ 0.4} & 89.3 {\tiny $\pm$ 0.4} & \textbf{92.1} {\tiny $\pm$ 0.3} & 93.0 {\tiny $\pm$ 0.1} & 93.3 {\tiny $\pm$ 0.4} \\
AL & 81.0 {\tiny $\pm$ 0.6} & 88.5 {\tiny $\pm$ 0.5} & 91.5 {\tiny $\pm$ 0.3} & 93.0 {\tiny $\pm$ 0.2} & 93.4 {\tiny $\pm$ 0.3} \\ \midrule
\end{tabular}}
  \caption{CIFAR-10 with GLISTER}
\end{table}

\subsection{Effect of Query Size}
\label{sec:query_size_effect}
In this section, we demonstrate that speedups with CAL methods can be realized with different query sizes. We demonstrate this by comparing AL with CAL-SDS2 on CIFAR-10, with three different choices of query sizes. We use entropy based uncertainty sampling as the acquisition function. We observe that CAL-SDS2 achieves comparable performance to standard AL with different query sizes, and the speedup can increase when the query size is reduced.

\begin{table}[H]
\centering
\scalebox{.9}{
\begin{tabular}{c|c|ccccc|ccccc}
\toprule[1.5pt]
\multicolumn{2}{c|}{}                                               & \multicolumn{5}{c|}{\textbf{Test Accuracy (\%)}}                                   & \multicolumn{5}{c}{\textbf{Factor Speedup}}                                 \\ \midrule 
                                                \textbf{Query Size} & \textbf{Method}    & \textbf{10\%} & \textbf{20\%} & \textbf{30\%} & \textbf{40\%} & \textbf{50\%} & \textbf{10\%} & \textbf{20\%} & \textbf{30\%} & \textbf{40\%} & \textbf{50\%} \\ \midrule
\multicolumn{1}{c|}{\multirow{2}{*}{5000}} & AL       & 79.1          & 88.6          & 91.7          & 93.5          & 93.8          & 1.0x          & 1.0x          & 1.0x          & 1.0x          & 1.0x          \\
\multicolumn{1}{c|}{}                           & \textcolor{blue}{CAL-SDS2} & 79.1          & 89.4          & 92.4          & \textbf{94.0} & 94.3          & 1.0x          & 1.1x          & 1.4x          & 1.7x          & 2.0x          \\ \midrule 
\multicolumn{1}{c|}{\multirow{2}{*}{2500}} & AL       & 81.9          & 89.6          & 92.4          & 93.5          & 94.1          & 1.0x          & 1.0x          & 1.0x          & 1.0x          & 1.0x          \\
\multicolumn{1}{c|}{}                           & \textcolor{blue}{CAL-SDS2} & 82.5          & 90.1          & 92.9          & 94.0          & \textbf{94.4} & 1.1x          & 1.6x          & 2.1x          & 2.7x          & 3.4x          \\ \midrule
\multicolumn{1}{c|}{\multirow{2}{*}{1000}} & AL    & 84.4          & \textbf{90.4} & \textbf{93.2} & 93.8          & 94.2          & 1.0x          & 1.0x          & 1.0x          & 1.0x          & 1.0x          \\
\multicolumn{1}{c|}{}                           & \textcolor{blue}{CAL-SDS2} & \textbf{84.3} & 89.8          & 92.5          & 93.2          & 93.5          & 1.6x          & 3.1x          & 4.2x          & 5.3x          & 6.4x          \\ \midrule 
\end{tabular}}
\caption{Effect of Query Size on CIFAR-10}
\end{table}

\section{Additional Details on Single-Cell Cell-Type Identity Classification Dataset}
\begin{figure}[H]
    \centering  
    \includegraphics[width=1\textwidth]{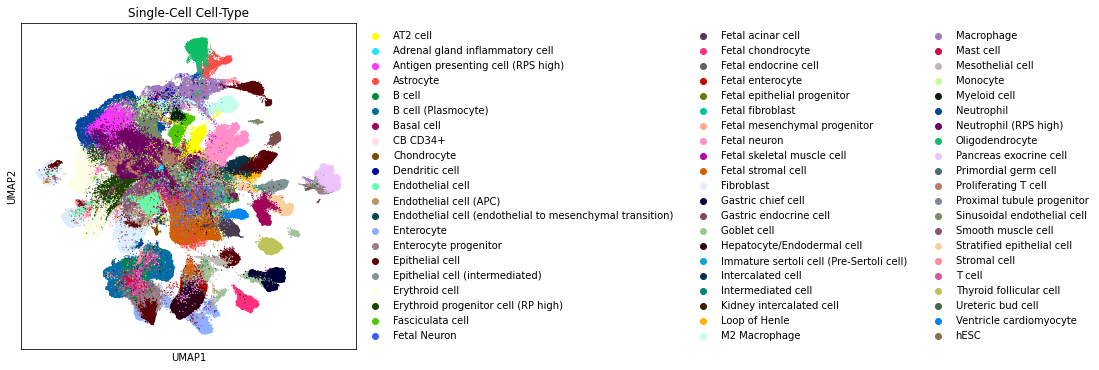} 
    \caption{UMAP embedding of single cells in HCL annotated by their cell type.}
    \label{fig:sc_umap}
\end{figure}

The human cell landscape (HCL) dataset consists of scRNA-seq data for 562,977 cells across 63 cell types represented in 56 human tissues. Each cell type may be present in multiple tissues. The cell type classes are highly imbalanced, with the rarest cell type, human embryonic stem cell, accounting for 0.06 \% of the total dataset and the most common, fibroblast, accounting for 6\%. The raw data is first normalized for library size and scaled to 10000 reads in total, followed by log transformation. We visualize the dataset using UMAP \ref{fig:sc_umap}.

\end{document}